\documentclass[times, 5p]{elsarticle}
\usepackage{hyperref}
\journal{-}

\usepackage{amsmath}
\usepackage{diagbox}
\usepackage{subfig}
\usepackage{pythonhighlight}

\usepackage{comment}
\newsavebox{\measurebox}

\def \resuneta {\texttt{ResUNet-a} }
\def \resunetamtsk {\texttt{ResUNet-a-mtsk} }
\def \resunetacmtsk {\texttt{ResUNet-a-cmtsk} }

\newcommand{\ie}{\textit{i.e.}, }
\newcommand{\eg}{\textit{e.g.}, }

\newcommand\numberthis{\addtocounter{equation}{1}\tag{\theequation}}

\hypersetup{
    colorlinks,
    linkcolor={red!50!black},
    citecolor={blue!50!black},
    urlcolor={blue!50!black}
}

\biboptions{authoryear}

\begin{document}

\begin{frontmatter}

\title{\texttt{ResUNet-a}: a deep learning framework for semantic segmentation of remotely sensed data}

\author[data61]{Foivos I. Diakogiannis\fnref{myfootnote1}}
\author[CSIROAF]{Fran\c{c}ois Waldner}
\author[data61]{Peter Caccetta}
\author[icrar]{Chen Wu}

\address[data61]{Data61, CSIRO, Floreat WA}
\fntext[myfootnote1]{foivos.diakogiannis@data61.csiro.au}
\address[CSIROAF]{CSIRO Agriculture \& Food, St Lucia, QLD, Australia}
\address[icrar]{ICRAR, The University of Western Australia, Crawley, WA}

\begin{abstract}
Scene understanding of high resolution aerial images is of great importance for the task of automated monitoring in various remote sensing applications. Due to the large within-class and small between-class variance in pixel values of objects of interest, this remains a challenging task. 
In recent years, deep convolutional neural networks  have started being used in remote sensing applications and demonstrate state of the art performance for pixel level classification of objects. 
\textcolor{black}{Here we propose a reliable framework for performant results for the task of semantic segmentation of monotemporal very high resolution aerial images. Our framework consists of a novel deep learning architecture, \texttt{ResUNet-a}, and a novel loss function based on the Dice loss.  
\texttt{ResUNet-a} uses a UNet encoder/decoder backbone, in combination with residual connections, atrous convolutions, pyramid scene parsing pooling and  multi-tasking inference. \texttt{ResUNet-a} infers sequentially the boundary of the objects, the distance transform of the segmentation mask, the segmentation mask and a colored reconstruction of the input. Each of the tasks is conditioned on the inference of the previous ones, thus establishing a conditioned relationship between the various tasks, as this is described through the architecture's computation graph. 
We  analyse the performance of several flavours of the Generalized Dice loss  for semantic segmentation, and we introduce a novel variant loss function for semantic segmentation of objects that has excellent convergence properties and behaves well even under the presence of highly imbalanced classes.} The performance of our modeling framework is evaluated on the ISPRS 2D Potsdam dataset. 
Results show state-of-the-art performance  with an average F1 score of 92.9\% over all classes for our best model.  
\end{abstract}


\end{frontmatter}

\section{Introduction}

Semantic labelling of very high resolution (VHR) remotely-sensed images, \ie the task of assigning a category to every pixel in an image, is of great interest for a wide range of 
urban applications including land-use planning, infrastructure  management, as well as urban sprawl detection~\citep{matikainen2011segment, zhang2011mapping, lu2017joint, goldblatt2018using}. Labelling tasks generally  focus on extracting one specific category, \eg  building, road, or certain vegetation types~\citep{li2015robust, cheng2017automatic, wen2017semanticaa}, or multiple classes all together~\citep{paisitkriangkrai2016semantic, langkvist2016classification, liu2018semantic, marmanis2018classification}.

Extracting spatially consistent information in urban environments from remotely-sensed imagery remains particularly challenging for two main reasons. First, urban classes often display a high within-class variability and a low between-class variability. On the one hand, man-made objects of the same semantic class are often built in different materials and with different structures, leading to an incredible diversity of colors, sizes, shapes, and textures. On the other hand, semantically-different man-made objects can present similar characteristics, \eg cement rooftops, cement sidewalks, and cement  roads. Therefore, objects with similar spectral signatures can belong to completely different classes.  Second, the intricate three-dimensional structure of urban environments is favourable to interactions between these objects, \eg through occlusions and cast shadows.  

Circumventing these issues requires going beyond the sole use of spectral information and including geometric elements of the urban class appearance such as pattern, shape, size, context, and orientation. Nonetheless, pixel-based classifications still fail to satisfy the accuracy requirements because they are affected by the salt-and-pepper effect and cannot fully exploit the rich information content of VHR data~\citep{myint2011per, li2014object}. GEographic Object-Based Imagery Analysis (GEOBIA) is an alternative image processing approach that seeks to group pixels into  meaningful objects based on specified parameters~\citep{blaschke2014geographic}. Popular image segmentation algorithm in remote sensing include watershed segmentation~\citep{vincent1991watersheds}, multi-resolution segmentation~\citep{baatz} and mean-shift segmentation~\citep{comaniciu2002mean}. In addition, GEOBIA also allows to compute additional attributes related to the texture, context, and shape of the objects, which can then be added to the classification feature set. However, there is no universally-accepted method to identify the segmentation parameters that provide optimal pixel grouping, which implies the GEOBIA is still highly interactive and includes subjective trial-and-error methods and arbitrary decisions. Furthermore, image segmentation  might fail to simultaneously address the wide range of object sizes that one typically encounters in urban landscapes ranging from finely structure objects such as cars and trees to larger objects such as buildings. Another drawback is that GEOBIA relies on pre-selected features for which the maximum attainable accuracy is \textit{a priori} unknown. While several methods have been devised to extract and select features, these methods are not themselves learned from the data, and are thus potentially sub-optimal.

In recent years, deep learning methods and Convolutional Neural Networks (CNNs) in particular \citep{DBLP:journals/neco/LeCunBDHHHJ89}  have surpassed traditional methods in various computer vision tasks, such as object detection, semantic, and instance segmentation
\citep[see][for a comprehensive review]{doi:10.1162/necoa00990}.  Some of the key advantages of CNN-based algorithms is that they provide end-to-end solutions, that require minimal feature engineering which offer greater generalization capabilities. They also perform object-based classification, \ie they take into account features that characterize entire image objects, thereby reducing the salt-and-pepper effect that affects conventional classifiers.

Our approach to  annotate image pixels with class labels is object-based, that is, the algorithm extracts characteristic features from whole (or parts of) objects that exist in  images such as cars, trees, or corners of buildings and assigns a vector of class probabilities to each pixel.  
In contrast, using   standard classifiers such as random forests,  the probability of each class per pixel is based on features inherent in the spectral signature only. Features based on spectral signatures contain less information than features based on objects. For example, looking at a car we understand not only it's spectral features (color) but also how these vary as well as the extent these occupy in an image. In addition, we understand that it is more probable a car to be surrounded by pixels belonging to a road, and less probable to be surrounded by pixels belonging to buildings.
In the field of computer vision, there is a vast literature on various modules used in convolutional neural networks that make use of this idea of ``per object classification''. These modules,   such as atrous convolutions \citep{DBLP:journals/corr/ChenPK0Y16} and pyramid pooling \citep{DBLP:journals/corr/HeZR014,zhao2017pspnet}, boost the algorithmic performance on semantic segmentation tasks. In addition, after the residual networks era \citep{DBLP:journals/corr/HeZRS15} it is now possible to train deeper neural networks avoiding to a great extent the problem of vanishing (or exploding) gradients.

Here, we introduce  a novel Fully Convolutional Network (FCN) for semantic segmentation, termed \resuneta. This network combines ideas distilled from computer vision applications of deep learning, and demonstrates competitive performance. In addition, we   describe a modeling framework consisting of  a new loss function that behaves well for semantic segmentation problems with class imbalance as well as for regression problems.  
In summary, the main contributions of this paper are the following:
\begin{enumerate}
\item A novel architecture for understanding and labeling very high resolution images \textcolor{black}{for the task of semantic segmentation.} The architecture
\textcolor{black}{uses a UNet \citep{DBLP:journals/corr/RonnebergerFB15} encoder/decoder backbone, in combination with, residual connections \citep{DBLP:journals/corr/HeZR016},  atrous convolutions \citep{DBLP:journals/corr/ChenPK0Y16,DBLP:journals/corr/ChenPSA17}, pyramid scene parsing pooling \citep{zhao2017pspnet} and  multi tasking inference \citep[][we present two variants of the basic architecture, a single task and a multi-task one]{DBLP:journals/corr/Ruder17a}.}           
\item We analyze the performance of various flavours of the Dice coefficient for semantic segmentation. Based on our findings, we introduce a variant of the Dice loss function  that speeds up the convergence of semantic segmentation tasks and improves  performance. Our results indicate that the new loss function behaves well even when there is a large class imbalance. This loss can also be used for continuous variables when the target domain of values is in the range [0,1].  
\end{enumerate}
\textcolor{black}{In addition, we also present a data augmentation methodology, where the input is viewed in multiple scales during training by the algorithm, that improves performance and avoids overfitting.}
\noindent  The performance of \resuneta was tested using the Potsdam data set made available through the ISPRS competition \citep{ISPRS}. Validation results show that \resuneta achieves state-of-the-art results.

This article is organized as follows. 
In section \ref{related_work} we provide a short review of related work on the topic of semantic segmentation focused on the field of remote sensing. In section \ref{resuneta}, we detail the model architecture and the modeling framework. Section \ref{resuneta_data} describes the data set we used for training our algorithm.   In section \ref{ablation_study_all} we provide an experimental analysis that justifies the design choices for our modeling framework. 
Finally, section \ref{resuneta_results} presents the performance evaluation of our algorithm and comparison with other published results. Readers are referred to sections \ref{resuneta_babysitting} for a description of our software implementation and hardware configurations, and to section \ref{resuneta_inf_results} for the full error maps on unseen test data.

\section{Related Work}
\label{related_work}

The task of semantic segmentation has attracted significant interest in the latest years, not only  in the field of computer vision community but also in other disciplines (e.g. biomedical imaging, remote sensing) where automated annotation of images is an important process. In particular, specialized techniques have been developed over different disciplines, since there are task-specific peculiarities that the community of computer vision does not have to address (and vice versa).  

Starting from the computer vision community, when first introduced, Fully Convolutional Networks (hereafter FCN) for semantic segmentation \citep{DBLP:journals/corr/LongSD14}, improved the state of the art by a significant margin (20\% relative improvement over the state of the art on the  PASCAL VOC \citep{Everingham10}  2011 and 2012 test sets). The authors replaced the last fully connected layers with convolutional layers. The original resolution was achieved with a combination of upsampling and skip connections. 
Additional improvements have been presented with the use of deeplab models \citep{DBLP:journals/corr/ChenPK0Y16,DBLP:journals/corr/ChenPSA17}, that first showcased the importance of atrous convolutions for the task of semantic segmentation. Their model uses also a conditioned random field as a post processing step in order to refine the final segmentation. 
A significant contribution in the field of computer vision came from the community of biomedical imaging and in particular, the U-Net architecture \citep{DBLP:journals/corr/RonnebergerFB15} that introduced the encoder-decoder paradigm, for upsampling gradually from lower size features to the original image size.  
Currently, the state of the art on the computer vision datasets is considered to be mask-rcnn \citep{DBLP:journals/corr/HeGDG17}, that performs various tasks (object localization, semantic segmentation, instance segmentation, pose estimation etc). A key element of the success of this architecture is its multitasking nature.

\textcolor{black}{
One of the major advantages of CNNs over traditional classification methods (e.g. random forests), is their ability to process input data in multiple context levels. This is achieved through  the  downsampling operations  that summarizes features. However, this advantage in feature extraction needs to be matched with a proper upsampling method, to retain information from all spatial resolution contexts and produce fine boundary layers. 
There has been a quick uptake of the approach in the remote sensing community and various solutions based on deep learning have been presented recently  
\textcolor{black}{\citep[e.g.][]{DBLP:journals/corr/Sherrah16,DBLP:journals/corr/AudebertSL16a,rs9040368,audebert2018beyond,
langkvist2016classification,li2015robust,li2014object,
volpi2017dense,liu2018semantic,rs9060522,liu2017dense,
rs10050743,s18113774,Marmanis2016SEMANTICSO,
marmanis2018classification,wen2017semanticaa,
ZHAO201748}. A comprehensive review of deep learning applications in the field of remote sensing can be found in
\citet{8113128,MA2019166,app9102110}}}.

\textcolor{black}{Discussing in more detail some of the most relevant approaches to our work, 
\citep{DBLP:journals/corr/Sherrah16} utilized the FCN architecture, with a novel no-downsampling approach based on atrous convolutions to mitigate this problem. The summary pooling operation was traded with atrous convolutions, for filter processing at different scales.  The best performing architectures from their experiments were the ones using  pretrained convolution networks. The loss used was categorical cross-entropy.} 

\cite{rs9060522} introduced the Hourglass-shape network for semantic segmentation on VHR images, which included an encoder-decoder style network, utilizing inception like modules. \textcolor{black}{Their encoder-decoder style departed from the UNet backbone, in that they did not use features from all spatial contexts of the encoder in the decoder branch. Also, their decoder branch is not symmetric to the encoder. The building blocks of the encoder are inception modules. Feature upsampling takes place with the use of transpose convolutions. The loss used was weighted binary cross entropy.}  

Emphasizing on the importance of using the information from the boundaries of objects, 
\textcolor{black}{\cite{marmanis2018classification} 
utilized the Holistically Ne-sted Edge Detection network \citep[][HED]{DBLP:journals/corr/XieT15} for predicting boundaries of objects.  The loss used for the boundaries was an Euclidean distance regression loss.  
The estimated boundaries were then concatenated with image features and provided them as input into another CNN segmentation network, for the final classification of pixels. For the CNN segmentation network, they experimented with two architectures, the SegNet \citep{DBLP:journals/corr/BadrinarayananK15} and a Fully Convolutional Network presented in \citet{Marmanis2016SEMANTICSO} that uses weights from pretrained architectures.   One of the key differences in our approach for boundary detection with  \citet{marmanis2018classification}, is that the boundary prediction happens at the end of our architecture, therefore the request for boundary prediction affects all features since the boundaries are strongly correlated with the extent of the predicted classes. In contrast, in \cite{marmanis2018classification}, the boundaries are fed as input to the segmentation branch of their network, i.e. the segmentation part of their network uses them as additional input. Another difference is that we do not use weights from pretrained networks.} 

\textcolor{black}{\cite{s18113774} presented the Dense Pyramid Network. 
The authors incorporated group convolutions to process independently the Digital Surface Model from the true orthophoto, presenting an interesting data fusion approach. The channels created from their initial group convolutions were shuffled, in order to enhance the information flow between channels.
The authors, utilized a DenseNet \citep{DBLP:journals/corr/HuangLW16a} architecture as their feature extractor. In addition, a Pyramid Pooling layer was used at the end of their encoder branch, before constructing the final segmentation classes. 
 In order to overcome the class imbalance problem, they chose to use the Focal loss function \citep{DBLP:journals/corr/abs-1708-02002}. In comparison with our work, the authors did not use a symmetric 
 encoder-decoder architecture. The building blocks of their model were DenseNet units  which are known to be more efficient than standard residual units \citep{DBLP:journals/corr/HuangLW16a}. The pyramid pooling operator used in the end of their architecture, before the final segmentation map, is at different scales than the one used in \texttt{ResUNet-a}.}
 
\textcolor{black}{\cite{liu2018semantic} introduced the CASIA network, which consists of a pretrained deep encoder, a set of self-cascaded convolutional units and a decoder part. The encoder part is deeper than the decoder part.  The upscaling of the lower level features takes place with a resize operation followed by a convolutional residual correction term. The self-cascaded units, consist of a sequential multi-context aggregation layer, that aggregates features from higher receptive fields to local receptive fields. In a similar idea to our approach, the CASIA network uses features from multiple contexts, however these are evaluated at a different depth of the network and fused together in a completely different way. 
The architecture achieved state of the art performance on the ISPRS Potsdam and Vaihingen data. The loss function they used was the normalized cross entropy.}

\section{The \resuneta framework}
\label{resuneta}

In this section, we introduce the architecture of \resuneta in full detail (section \ref{resuneta_full_arch}), a novel loss function design to achieve faster convergence and higher performance (section \ref{resuneta_loss}),  data augmentation methodology (section \ref{resuneta_data_augmentation}) as well as the methodology we followed on performing inference on large images (section \ref{resuneta_inference}). The training strategy and software implementation characteristics  can be found in \ref{resuneta_babysitting}.

\begin{figure*}[h!!!]
\centering
\sbox{\measurebox}{%
  \begin{minipage}[b]{.5\textwidth}
  \subfloat
    []
    {\label{subfig:resuneta_full}\includegraphics[width=\textwidth]{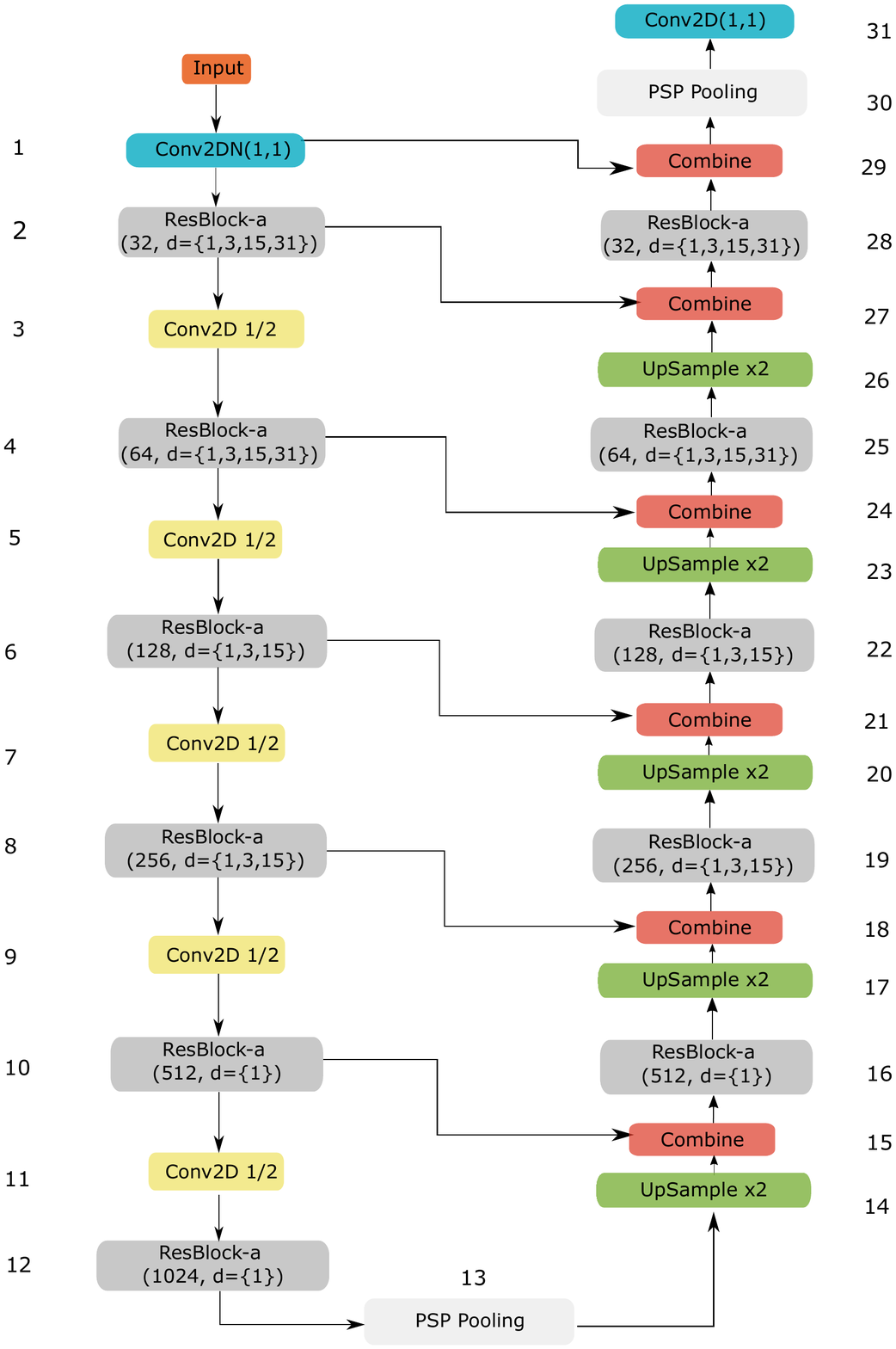}}
  \end{minipage}}

\usebox{\measurebox}\qquad
\begin{minipage}[b][\ht\measurebox][s]{.425\textwidth}
\centering
\subfloat
  []
  {\label{subfig:resuneta_bblock}\includegraphics[width=0.85\textwidth]{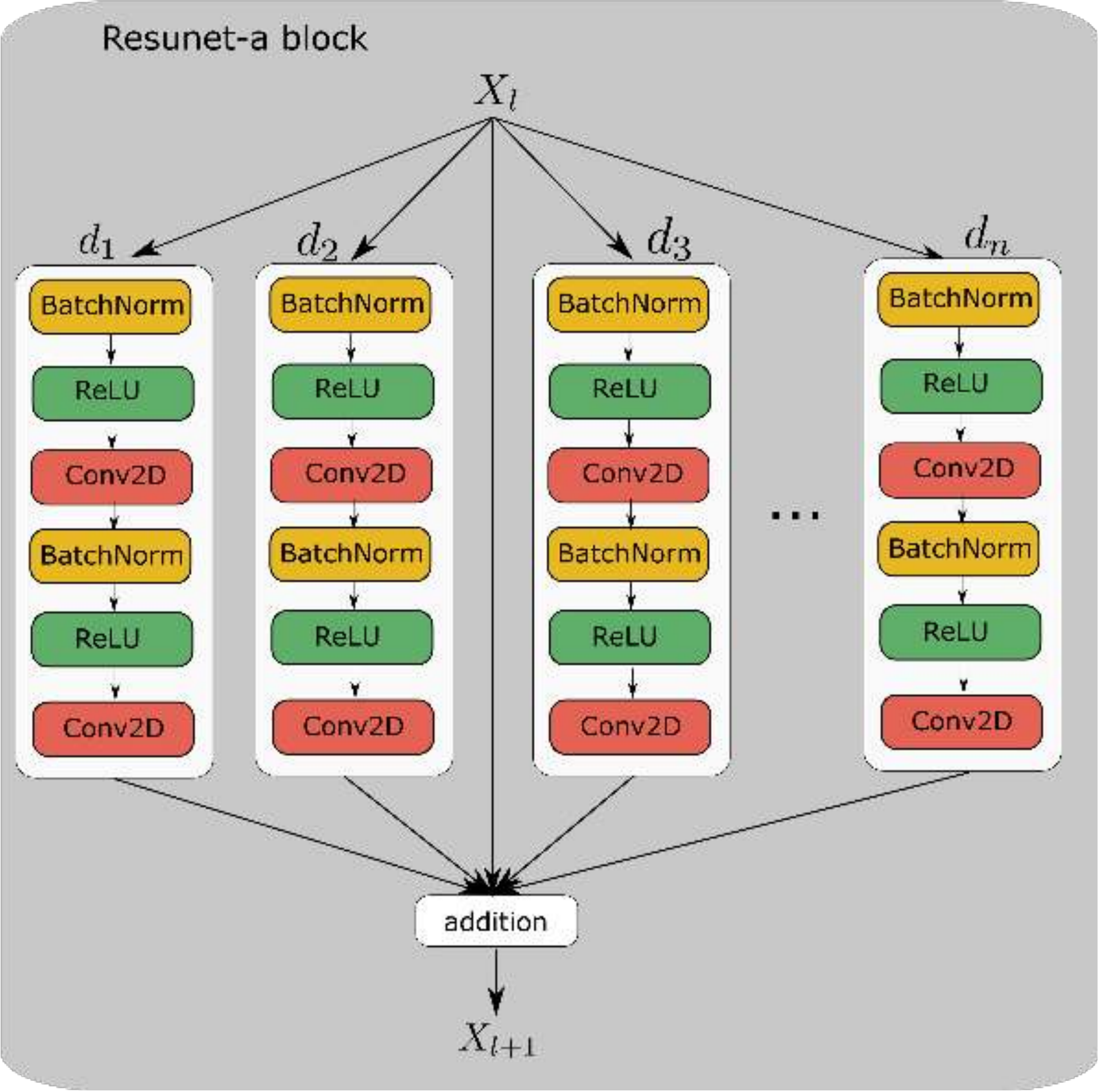}}
\vfill
\subfloat
  []
  {\label{subfig:resuneta_pspooling}\includegraphics[width=0.85\textwidth]{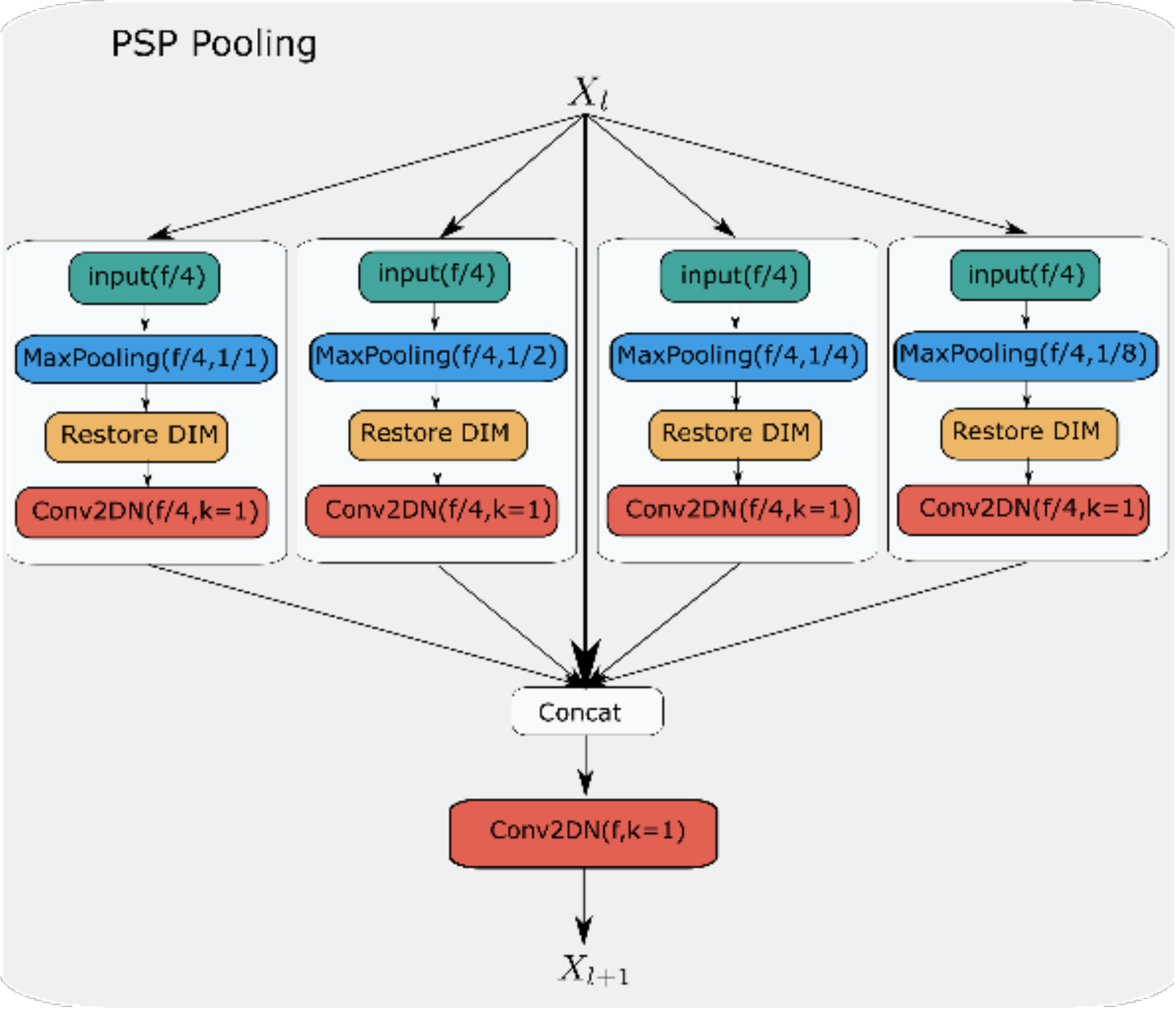}}
\end{minipage}
\caption{Overview of the \resuneta  d6 network. (a) The left (downward) branch is the encoder part of the architecture. The right (upward) branch is the decoder. The last convolutional layer has as many channels as there are distinct classes. (b) Building block of the \resuneta  ~network. Each unit within the residual block has the same number of filters with all other units. \textcolor{black}{Here $d_1,\ldots,d_n$ designate different dilation rates}, (c) Pyramid scene parsing pooling layer. \textcolor{black}{Pooling takes place in 1/1, 1/2, 1/4 and 1/8 portions of the original image.}}
\label{fig:resuneta}
\end{figure*}

\begin{figure}
\centering
\includegraphics[width=\columnwidth]{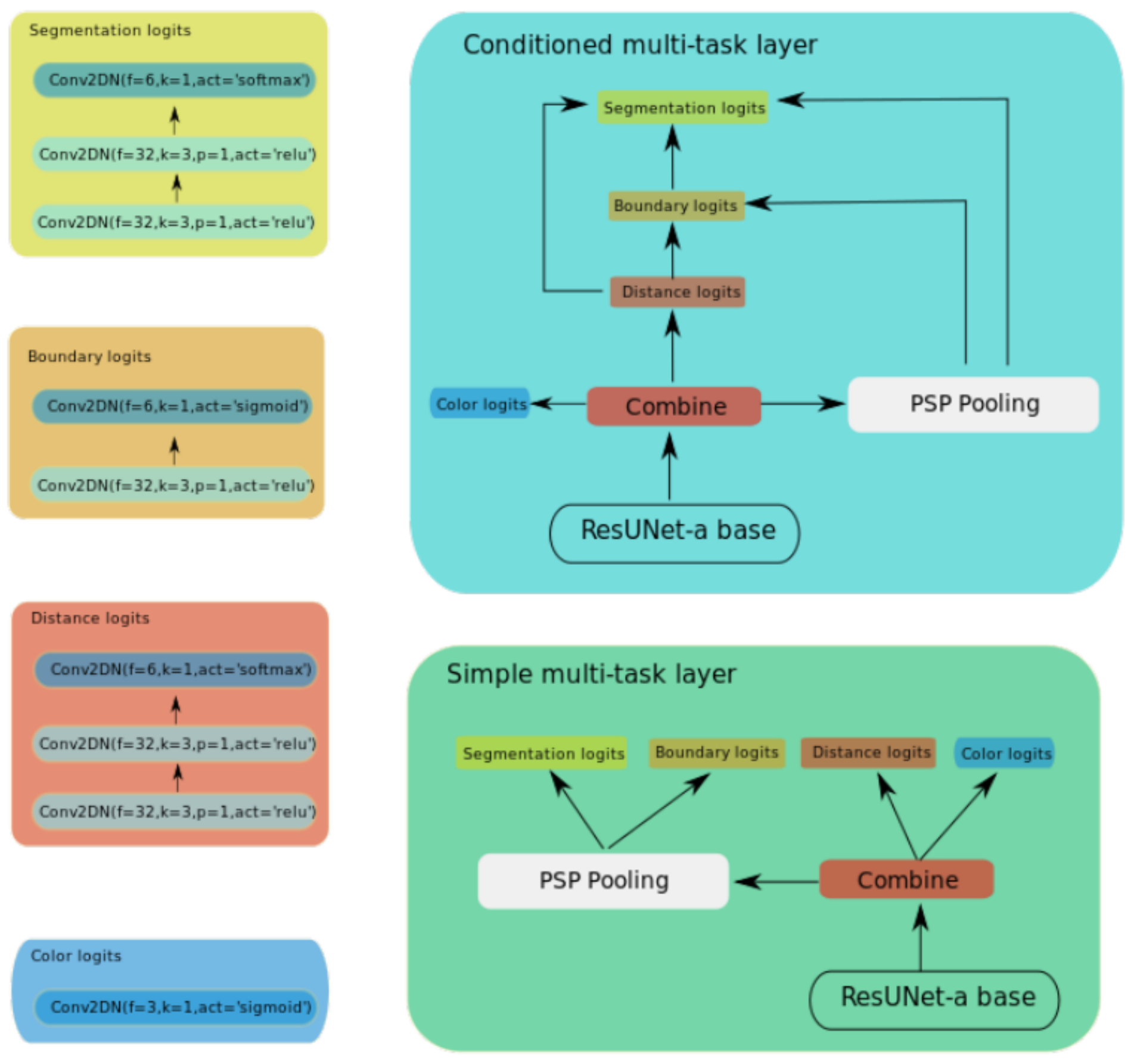}
\caption{Multi-task output layer of the \resuneta network. Referring to the example of \resuneta d6,  Layers 29, 30 and 31 are replaced with one of two variants of the multi-task layer. The first one, the  \textcolor{black}{conditioned} multitask layer, combines the various intermediate products progressively 
so as the final segmentation layer to take a ``decision'' based on inference from previous results. The simple multi-task layer keeps the tasks independent. }
\label{mtsklayers}
\end{figure}

\subsection{Architecture}
\label{resuneta_full_arch}

Our architecture combines the following set of modules encoded in our models:  
\begin{enumerate}
\item  A UNet \citep{DBLP:journals/corr/RonnebergerFB15} backbone architecture, \ie the encoder-decoder paradigm, is selected for smooth and gradual transitions from the image to the segmentation mask. 
\item To achieve consistent training as the depth of the network increases, the building blocks of the UNet architecture were replaced with modified residual blocks of convolutional layers~\citep{DBLP:journals/corr/HeZR016}. 
Residual blocks  remove to a great extent the problem of vanishing and exploding gradients that is present in deep architectures.  

\item For better understanding across scales, multiple parallel atrous convolutions \citep{DBLP:journals/corr/ChenPK0Y16,DBLP:journals/corr/ChenPSA17} with different dilation rates are employed within each residual building block. Although it is not completely clear why atrous convolutions perform well, the intuition behind their usage is that they increase the receptive field of each layer.  The rationale of using these multiple-scale layers is to extract object features 
at various receptive field scales. The hope is that this will improve performance by identifying correlations between objects at different locations in the image. 

\item In order to enhance the performance of the network by including  background context information we use  the pyramid scene parsing pooling \citep{zhao2017pspnet} layer. In shallow architectures, where the last layer of the encoder has a size no less than 16x16 pixels, we use  this layer in two locations within the architecture: after the encoder part (\ie middle of the network) and the second last layer before the creation of the segmentation mask.  For deeper architectures, we use this layer only close to the last output layer.   

\item In addition to the standard architecture that has a single segmentation mask layer as output, we also present two models where we perform multi-task learning. The algorithm learns simultaneously four complementary tasks. The first is the segmentation mask. The second is  the common boundary between the segmentation masks that is known to improve performance for semantic segmentation \citep{DBLP:journals/corr/BertasiusST15a,marmanis2018classification}. The third is the distance transform\footnote{The result of the distance transform on a binary segmentation mask is a gray level image, that takes values in the range [0,1], where each pixel value corresponds to the distance to the closest boundary. In \textsc{OpenCV} this transform is encoded in \texttt{cv::distance\_transform}.} 
\citep{Borgefors:1986:DTD:17140.17147} of the segmentation mask. The fourth is the actual colored image, in HSV color space. That is, the identity transform of the content, but in a different color space.     
\end{enumerate}

We term our network \resuneta ~because it consists of residual building blocks with multiple atrous convolutions and a UNet backbone architecture. 
We present two  basic architectures, \resuneta d6 and \resuneta d7, that differ in their depth, i.e. the total number of layers. In \resuneta d6  the encoder part consists of six \texttt{ResBlock-a} building blocks followed by a \texttt{PSPPooling} layer. In \resuneta d7 the encoder consists of seven \texttt{ResBlock-a} building blocks.  For each of the d6 or d7 models, there are also three different output possibilities: a  single task semantic segmentation layer,  a multi-task layer (mtsk), and a  \textcolor{black}{conditioned} multi-task output layer (cmtsk). The difference between the mtsk and cmtsk output layers is how the various complementary tasks (i.e. the boundary, the distance map, and the color) are used for the determination of the main target task, which is the  semantic segmentation prediction.   
 In the following we present in detail these models, starting from the basic \resuneta d6.

\subsubsection{\resuneta }

The \resuneta d6 network consists of stacked layers of modified residual building blocks (\texttt{ResBlock-a}), in an encoder-decoder style (UNet). 
The input is initially subjected to a  convolution layer of kernel size $(1,1)$ to increase the number of features to the desired initial filter size. A $(1,1)$ convolution layer was used in order to avoid any information loss from the initial image by summarizing features across pixels with a larger kernel. Then follow the residual blocks. 
In each residual block (Fig. \ref{subfig:resuneta_bblock}),  we used as many as three in parallel atrous convolutions in addition to the standard set of two convolutions of the residual network architecture, \ie there were up to four parallel branches of sets of two stacked convolutional layers.  After the convolutions, the output is added to the initial input in the spirit of residual building blocks.  We decided to sum the various atrous branches (instead of concatenating them) because it is known that the residual blocks of two successive convolutional layers demonstrate constant condition number of the Hessian of the loss function, irrespective of the depth of the network  \cite{DBLP:journals/corr/LiJHW16}. Therefore the summation scheme is easier to train (in comparison with the concatenation of features).  
In the encoder part of the network, the output of each of the residual blocks is downsampled with a convolution of kernel size of one and stride of two.  At the end of both the encoder and the decoder part, there exists a \texttt{PSPooling} operator \citep{zhao2017pspnet}. 
In the \texttt{PSPPooling} operator (Fig. \ref{subfig:resuneta_pspooling}), the initial input is split in channel (feature) space in 4 equal partitions. Then we perform max pooling operation in successive splits of the input layer, in 1, 4, 16 and 64 partitions.  Note that in the middle layer (Layer 13 has size: [batch size]$\times 1024 \times 8 \times 8$), the split of 64 corresponds to the actual total size of the input (so we have no additional gain with respect to max pooling from the last split). In Fig. \ref{subfig:resuneta_full} we present the full architecture of \resuneta ~(see also Table \ref{ResUNeta_layers}).  
In the decoder part, the upsampling is being done with the use of nearest neighbours interpolation followed by a normed convolution with a kernel size of one. 
By normed convolution, denoted with \texttt{Conv2DN}, we mean  a set of a single 2D convolution followed by a \texttt{BatchNorm} layer.
This approach for increasing the resolution of the convolution features was used in order  to avoid the chequerboard artifact in the segmentation mask \citep{46191}. The combination of layers from the encoder and decoder parts is being performed with the \texttt{Combine} layer (Table \ref{ResUNeta_comb_layer}). This module concatenates the two inputs and subjects them to a normed convolution that brings the number of features to the desired size.  

\textcolor{black}{The \resuneta d7 model is deeper than the corresponding d6 model, by one resunet building block both in the encoder and decoder parts. We have tested two versions of this deeper architecture that differ in the way the pooling takes place in the middle of the network. 
In version 1 \textcolor{black}{(hereafter d7v1)} the \texttt{PSPPooling} Layer (\texttt{Layer} 13) is replaced with one additional building block, that is a standard resnet block (see Table \ref{ResUNeta_PSPPooling_Replacement} for details). 
There is, of course, a corresponding increase  in the layers of the decoder part as well, by one additional residual building block.  In more detail 
(Table \ref{ResUNeta_PSPPooling_Replacement}), 
the \texttt{PSPPooling} layer in the middle of the network is replaced by a standard residual block at a lower resolution. The output of this layer is subjected to a   \texttt{MaxPooling2D}(\texttt{kernel}=2, \texttt{stride}=2)  operation  the output of which is rescaled to its original size and then concatenated with the original input layer. This operation is followed by a standard convolution that brings the total number of features (i.e. the number of channels) to their original number before the concatenation.
In version 2 \textcolor{black}{(hereafter d7v2)}, again the \texttt{Layer 12} is replaced with a standard resnet block. However, now the \texttt{MaxPooling}  operation following this layer is replaced with a smaller \texttt{PSPPooling} layer that has three parallel branches, performing pooling in 1/1, 1/2, 1/4 scales of the original filter (Fig. \ref{subfig:resuneta_pspooling}).  The reason for this is that the filters  in the middle of the d7 network  cannot sustain 4 parallel pooling operations due to their small size (therefore, we remove the 1/8 scale pooling), for an initial input image of size 256x256.}

With regards to the model complexity, \resuneta d6 has $\sim$ 52M trainable parameters for an initial filter size of 32.  
\resuneta d7 that has greater depth has $\sim$ 160M parameters for the same initial filter size. The number of parameters remains almost identical for the case of the multi-task models as well.

\begin{table}
\footnotesize
\caption{Details of the \resuneta layers for the d6 model. Here \texttt{f} stands for the number of output channels (or features, the input number of features is deduced from the previous layers). \texttt{k} is the convolution kernel size, \texttt{d} is the dilation rate, and \texttt{s}  the stride of the convolution operation. In all convolution operations we used appropriate zero padding to keep the dimensions of the produced feature maps equal to the input feature map (unless downsampling).}
\label{ResUNeta_layers}
\begin{center}
\begin{tabular}{ | l | l | }\hline
Layer \# & Layer Type\\\hline\hline
 1 & \texttt{Conv2D}(\texttt{f}=32, \texttt{k}=1, \texttt{d}=1, \texttt{s}=1)\\\hline
2 & \texttt{ResBlock-a}(\texttt{f}=32, \texttt{k}=3, \texttt{d}=\{1,3,15,31\}, \texttt{s}=1)\\\hline
3 & \texttt{Conv2D}(\texttt{f}=64, \texttt{k}=1, \texttt{d}=1, \texttt{s}=2)\\\hline
4 & \texttt{ResBlock-a}(\texttt{f}=64, \texttt{k}=3, \texttt{d}=\{1,3,15,31\}, \texttt{s}=1)\\\hline
5 & \texttt{Conv2D}(\texttt{f}=128, \texttt{k}=1, \texttt{d}=1, \texttt{s}=2)\\\hline
6 & \texttt{ResBlock-a}(\texttt{f}=128, \texttt{k}=3, \texttt{d}=\{1,3,15\}, \texttt{s}=1)\\\hline
7 & \texttt{Conv2D}(\texttt{f}=256, \texttt{k}=1, \texttt{d}=1, \texttt{s}=2)\\\hline
8 & \texttt{ResBlock-a}(\texttt{f}=256, \texttt{k}=3, \texttt{d}=\{1,3,15\}, \texttt{s}=1)\\\hline
9 & \texttt{Conv2D}(\texttt{f}=512, \texttt{k}=1, \texttt{d}=1, \texttt{s}=2)\\\hline
10 & \texttt{ResBlock-a}(\texttt{f}=512, \texttt{k}=3, \texttt{d}=1, \texttt{s}=1)\\\hline
11 & \texttt{Conv2D}(\texttt{f}=1024, \texttt{k}=1, \texttt{d}=1, \texttt{s}=2)\\\hline
12 & \texttt{ResBlock-a}(\texttt{f}=1024, \texttt{k}=3, \texttt{d}=1, \texttt{s}=1)\\\hline
13 & \texttt{PSPPooling} \\ \hline
14 & \texttt{UpSample} (\texttt{f}=512) \\ \hline
15 & \texttt{Combine} (\texttt{f}=512, Layers 14 \& 10) \\ \hline
16 & \texttt{ResBlock-a}(\texttt{f}=512, \texttt{k}=3, \texttt{d}=1, \texttt{s}=1)\\\hline
17 & \texttt{UpSample} (\texttt{f}=256) \\ \hline
18 & \texttt{Combine} (\texttt{f}=256, Layers 17 \& 8) \\ \hline
19 & \texttt{ResBlock-a}(\texttt{f}=256, \texttt{k}=3, \texttt{d}=1, \texttt{s}=1)\\\hline
20 & \texttt{UpSample} (\texttt{f}=128) \\ \hline
21 & \texttt{Combine} (\texttt{f}=128, Layers 20 \& 6) \\ \hline
22 & \texttt{ResBlock-a}(\texttt{f}=128, \texttt{k}=3, \texttt{d}=1, \texttt{s}=1)\\\hline
23 & \texttt{UpSample} (\texttt{f}=64) \\ \hline
24 & \texttt{Combine} (\texttt{f}=64, Layers 23 \& 4) \\ \hline
25 & \texttt{ResBlock-a}(\texttt{f}=64, \texttt{k}=3, \texttt{d}=1, \texttt{s}=1)\\\hline
26 & \texttt{UpSample} (\texttt{f}=32) \\ \hline
27 & \texttt{Combine} (\texttt{f}=32, Layers 26 \& 2) \\ \hline
28 & \texttt{ResBlock-a}(\texttt{f}=32, \texttt{k}=3, \texttt{d}=1, \texttt{s}=1)\\\hline
29 & \texttt{Combine} (\texttt{f}=32, Layers 28 \& 1) \\ \hline
30 & \texttt{PSPPooling} \\ \hline
31 & \texttt{Conv2D} (\texttt{f} = NClasses, \texttt{k}=1, \texttt{d}=1, \texttt{s}=1)\\ \hline
32 & \texttt{Softmax}(\texttt{dim} = 1) \\\hline
\end{tabular}
\end{center}
\end{table}

\begin{table}
\footnotesize
\caption{Details of the \texttt{Combine}(\texttt{Input1},\texttt{Input2}) layer.}
\label{ResUNeta_comb_layer}
\begin{center}
\begin{tabular}{ | l | l | }\hline
Layer \# & Layer Type\\\hline\hline
 1 & \texttt{Input1}\\\hline
2 & \texttt{ReLU}(\texttt{Input1})\\\hline
3 & \texttt{Concat}(Layer 2,\texttt{Input2}) \\ \hline
3 & \texttt{Conv2DN}(\texttt{k}=1, \texttt{d}=1, \texttt{s}=1)\\\hline
\end{tabular}
\end{center}
\end{table}

\begin{table}
\footnotesize
\caption{Details of the replacement of the middle \texttt{PSPPooling} layer (\texttt{Layer} 13 from Table \ref{ResUNeta_layers}) for the \resuneta d7 model.}
\label{ResUNeta_PSPPooling_Replacement}
\begin{center}
\begin{tabular}{ | l | l | }\hline
Layer \# & Layer Type\\\hline\hline
 input & \texttt{Layer 12} \\\hline 
 A & \texttt{Conv2D}(\texttt{f}=2048, \texttt{k}=1, \texttt{d}=1, \texttt{s}=2)(input)\\\hline
B &  \texttt{ResBlock-a}(\texttt{f}=2048, \texttt{k}=3, \texttt{d}=1, \texttt{s}=1)(\texttt{Layer} A)\\\hline
C & \texttt{MaxPooling}(\texttt{kernel}=2, \texttt{stride}=2)(\texttt{Layer} B) \\ \hline
D & \texttt{UpSample}(\texttt{Layer} C)\\\hline
E & \texttt{Concat}(\texttt{Layer} D, \texttt{Layer} B)\\\hline
F & \texttt{Conv2D}(\texttt{f}=2048,\texttt{kernel}=1)(\texttt{Layer} E)\\\hline
\end{tabular}
\end{center}
\end{table}

\subsubsection{Multitasking \resuneta}
This version of \resuneta replaces the last layer (Layer 31) with a multitasking block (Fig. \ref{mtsklayers}). The multiple tasks are complementary. These are (a) the prediction of the semantic segmentation mask, (b) the detection of the common boundaries between classes, (c) the reconstruction of the distance map and (e) the reconstruction of the original image in HSV color space.  
\textcolor{black}{Our choice of using a different color space than the original input was guided by the principle that: (a) we wanted to avoid the identity transform in order to exclude the algorithm recovering trivial solutions and (b) the HSV (or HSL) colorspace matches closely the human perception of color \cite{A.Vadivel2005}.}
It is important to note that these additional labels are derived using standard computer vision libraries from the initial image and segmentation mask, without the need for additional information (e.g. separately annotated boundaries). A software implementation for this is given in \ref{bound_dist_estimate}. The idea here is that all these tasks are complementary and should help the target task that we are after. Indeed, the distance map provides information for the topological connectivity of the segmentation mask as well as the extent of the objects (for example if we have an image with a ``car'' (object class) on a ``road'' (another object class), then the ground truth of the mask of the ``road'' will have a hole exactly to the location of the pixels corresponding to the ``car'' object). The boundary  helps in better understanding the extent of the segmentation mask. Finally, the colorspace transformation provides additional information for the correlation between color variations and object extent. It also helps to keep ``alive'' the  information of the fine details of the original image to its full extent until the final output layer. The rationale here is similar with the idea behind the concatenation of higher order features (first layers) with lower order  features  that exist in the UNet backbone architecture: the encoder layers have finer details about the original image as closely as they are to the original input. Hence, the reason for  concatenating them  with the layers of the decoder is to keep the fine details necessary until the final layer of the network that is ultimately responsible for the creation of the segmentation mask. By demanding the network to be able to reconstruct the original image, we are making sure that all fine details are preserved\footnote{However, the color reconstruction on its own does not guarantee that the network learns meaningful correlations between classes and colors.}  \textcolor{black}{(an example of input image, ground truth and inference for all the tasks in the conditioned multitasking setting can be seen in Fig. \ref{resuneta_inference_all_demo}).}

We present two flavours of the algorithm whose main difference is how the various tasks are used for the target output that we are interested in. In the simple multi-task block (bottom right block of Fig \ref{mtsklayers}), the four tasks are produced simultaneously and independently. That is, there is no direct usage of the three complementary tasks (boundary, distance, and color) in the construction of the target task that is the segmentation.  The motivation here is that the different tasks will force the algorithm to identify new meaningful features that are correlated with the output we are interested in and can help in the performance of the algorithm for semantic segmentation. For the distance map, as well as the color reconstruction, we do not use the PSPPooling layer. This is because it tends to produce large squared areas with the same values (due to the pooling operation) and the depth of the convolution layers in the logits is not sufficient to diminish this. 

The second version of the algorithm uses a \textcolor{black}{conditioned} inference methodology. That is, the network graph is constructed in such a way so as to take advantage of the inference of the previous layers (top right block of Fig \ref{mtsklayers}). We first predict the distance map. The distance map is then concatenated with the output of the PSPPooling layer and is used to calculate the boundary logits. Then both the distance map and the prediction of the boundary are concatenated with the PSPPooling layer and the result is provided as input to the segmentation logits for the final prediction.

\subsection{Loss function}
\label{resuneta_loss}

In this section, we introduce a new variant of the family of Dice loss functions for semantic segmentation and regression problems.   \textcolor{black}{The Dice family of losses is by no means the only option for the task of semantic segmentation. Other interesting loss functions for the task of semantic segmentation are the focal loss \citealt[][see also \cite{s18113774} for an application on VHR images]{DBLP:journals/corr/abs-1708-02002},  the boundary loss \citep{kervadec2018boundary},  and the Focal Tversky loss \citep{DBLP:journals/corr/abs-1810-07842}. A list of many other available loss functions can be found in \citet{taghanaki2019deep}.}

\begin{figure*}[h!]
\includegraphics[width=\linewidth]{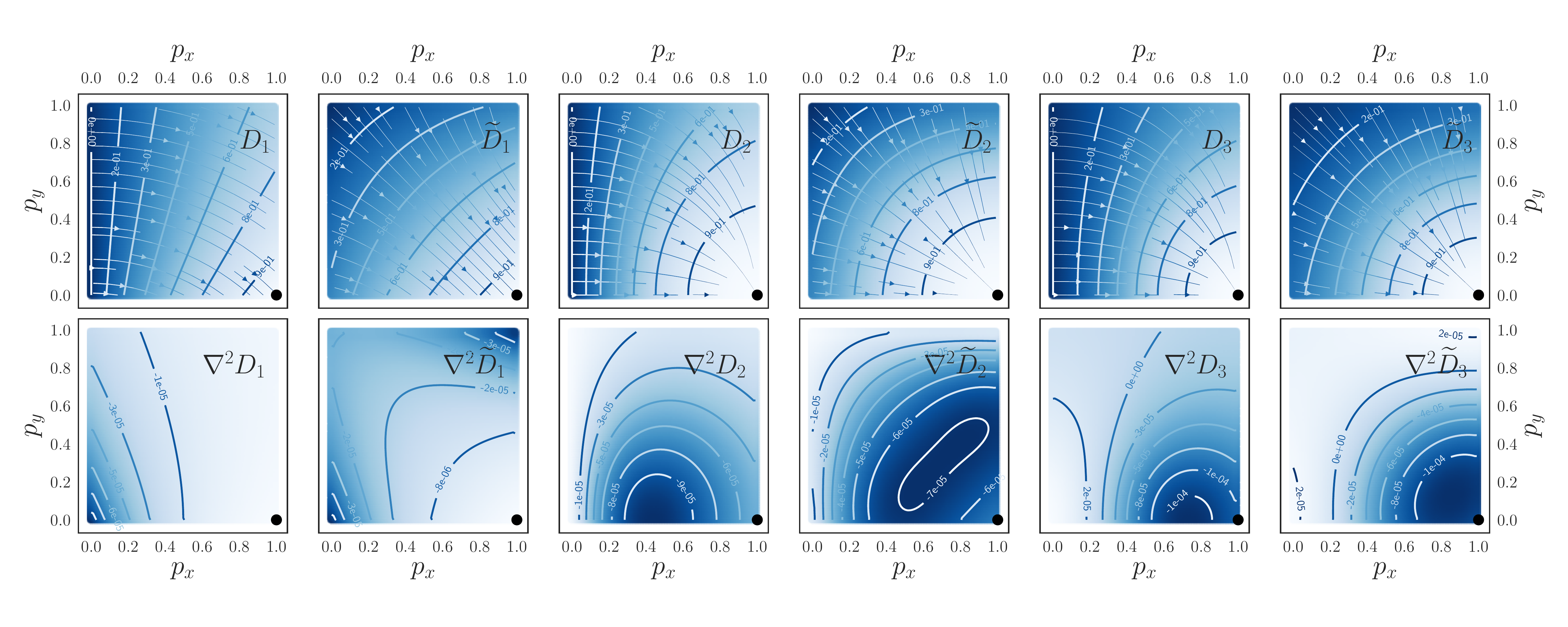}
    \caption{ \textcolor{black}{Contour plots of the gradient flow (top row) and Laplacian operator (bottom row) for the various versions of the Dice  loss functions (Eq \ref{eqn:dice1}--\ref{eqn:dice3}), $D_i$, as well as the functional forms with complements, $\tilde{D}_i$. The black dot corresponds to the ground truth value $(1,0)$.  From left to right, for the top row, we have the gradient flow of the generalized loss functions $D_1$, $\tilde{D}_1$, $D_2$,  $\tilde{D}_2$ and  $D_3$, $\tilde{D}_3$. The bottom panels are the corresponding Laplacian operators of these.  The numerical values of the isocontours on the images describe numerically the colorscheme with darker values corresponding to smaller values. }}
\label{ResUNetAAllLossnLaplacians}
\end{figure*}

\subsubsection{Introducing the Tanimoto loss with complement}
When it comes to semantic segmentation tasks, there are various options for the loss function. 
The Dice coefficient \citep{doi:10.2307/1932409,Sorensen-1948-BK}, generalized for fuzzy binary vectors in a multiclass context~\cite[][see also \citealt{journals/tmi/CrumCH06,DBLP:journals/corr/SudreLVOC17}]{DBLP:journals/corr/MilletariNA16}, is a popular choice among practitioners. 
It has been shown that it can increase performance over the cross entropy loss \citep{DBLP:journals/corr/NovikovMLHWB17}. 
The Dice coefficient can be generalized to continuous binary vectors in two ways: either by the summation of probabilities in the denominator or by the summation of their squared values. In the  literature, there are at least three definitions which are  equivalent \citep{journals/tmi/CrumCH06,DBLP:journals/corr/MilletariNA16,DBLP:journals/corr/DrozdzalVCKP16,DBLP:journals/corr/SudreLVOC17}:

\begin{align*}
D_1(\mathbf{p},\mathbf{l}) &= \frac{2\sum_i p_i l_i}{\sum_i p_i + \sum_i l_i} \numberthis \label{eqn:dice1} \\ 
D_2(\mathbf{p},\mathbf{l}) &= \frac{2\sum_i p_i l_i}{\sum_i (p_i^2 +  l_i^2)} \numberthis \label{eqn:dice2} \\ 
D_3(\mathbf{p},\mathbf{l}) &= \frac{\sum_i p_i l_i}{\sum_i (p_i^2 +  l_i^2) - \sum_i (p_i l_i)} \numberthis \label{eqn:dice3}\\
\end{align*} 
where $\mathbf{p} \equiv \{ p_i\}$, $p_i \in [0,1]$ is a continuous variable, representing the vector of probabilities for the $i$-th pixel, and $\mathbf{l}\equiv \{l_i\}$ are the corresponding ground truth labels. For binary vectors, $l_i \in \{0,1\}$. \textcolor{black}{In the following we will represent (where appropriate) for simplicity the set of vector coordinates, $\mathbf{p} \equiv\{ p_i\}$, with their corresponding tensor index notation, i.e  $\mathbf{p} \equiv\{ p_i\} \to p_i$.}

These three definitions are numerically equivalent, in the sense that they map the vectors $(p_i,l_i)$ to the continuous domain $[0,1]$, i.e.  $D(p_i,l_i):\Re^2 \to [0,1]$. The gradients however, of these loss functions behave differently for gradient based optimization, \ie for deep learning applications, as demonstrated in  \citet{DBLP:journals/corr/MilletariNA16}. In the remainder of this paper, we call Dice loss, the loss function with the functional form with the summation of probabilities and labels in the denominator (Eq.~\ref{eqn:dice1}). We also use the name Tanimoto for the $D_3$ loss function (Eq.~\ref{eqn:dice3}) and designate it with the letter $T\equiv D_3$.

We found empirically that the loss functions containing squares in the denominator behave better in pointing to the ground truth irrespective of the random initial configuration of weights. In addition, we found that we can achieve faster training convergence by complementing the loss with a dual form that measures the overlap area of the complement of the regions of interest.   That is, if $p_i$ measures the probability of the $i$th pixel to belong in class $l_i$, the complement loss is defined as $T(\mathbf{1}-p_i, \mathbf{1}-l_i)$, where the subtraction is performed element-wise, e.g. $\mathbf{1}-p_i=\{1-p_1, 1-p_2,\ldots, 1-p_n\}$ etc. 
The intuition behind the usage of the complement in the loss function comes from the fact that the numerator of the Dice coefficient, $\sum_i p_i l_i$, can be viewed as an inner product between the probability vector, $\mathbf{p} = \{ p_i\}$ and the ground truth label vector, $\mathbf{l}=\{ l_i\}$. Then, the part of the probabilities vector, $p_i$, that corresponds to the elements of the label vector, $l_i$, that have zero entries, does not alter the value of the inner product\footnote{As a simple example, consider four dimensional vectors, say $p = (p_1,p_2,p_3,p_4)$ and $l = (1,1,0,0)$. The value of the inner product term is $\mathbf{p}\cdot \mathbf{l} = p_1+p_2$, and therefore the information contained in $p_3$ and $p_4$ entries is not apparent to the numerator of the loss. The complement inner product provides information for these terms: $(\mathbf{1}-\mathbf{p}) \cdot (\mathbf{1} - \mathbf{l}) = (\mathbf{1}-\mathbf{p}) \cdot (0,0,1,1)  = 2-(p_3+p_4)$.}. 
We, therefore, propose that the best flow of gradients (hence faster training) is achieved using as a loss function the average of $T(p_i,l_i)$ with its complement, $T(\mathbf{1}-p_i, \mathbf{1}-l_i)$: 
\begin{equation}
\label{Tanimoto_wth_dual}
\tilde{T}(p_i,l_i) = \frac{T(p_i,l_i) + T(\mathbf{1}-p_i, \mathbf{1}-l_i)}{2}. 
\end{equation}

\subsubsection{Experimental comparison with other  Dice loss functions}
In order to justify these choices, we present an example with a single 2D ground truth vector, ($l=(1,0)$), and a vector of probabilities $p = (p_x,p_y)\in [0,1]^2$. We consider the following six loss functions:
\begin{enumerate}
    \item the Dice coefficient, $D_1(p_i,l_i)$ ((Eq. \ref{eqn:dice1}))     
    \item the Dice coefficient with its complement: 
    $$ \tilde{D}_1(p_i,l_i) = (D_1(p_i,l_i)+D_1(\mathbf{1}-p_i,\mathbf{1}-l_i))/2$$
    \item The Dice coefficient $D_2(p_i,l_i)$ (Eq. \ref{eqn:dice2}).  
    \item the Dice coefficient with its complement, $\tilde{D}_2$. 
    \item the Tanimoto coefficient, $T(p_i,l_i)$ (Eq. \ref{eqn:dice3}). 
    \item the Tanimoto coefficient with its complement, $\tilde{T}(p_i,l_i)$ (Eq. \ref{Tanimoto_wth_dual}). 
\end{enumerate}

\textcolor{black}{In Fig.~\ref{ResUNetAAllLossnLaplacians} we plot the gradient field of the various flavours of the family of Dice loss functions (top panels), as well as the Laplacians of these (i.e. their 2nd order derivatives, bottom panels).} The ground truth is marked with a black dot. What is important in these plots is that for a random  initialization of the weights for a neural network, the loss function will take a (random) value in the area within $[0,1]^2$. The quality of the loss function then, as a suitable criterion for training deep learning models, is whether the gradients, from \emph{every point of the area} in the plot, direct the solution towards the ground truth point. 
Intuitively we also expect that the behavior of the gradients is even better, if the local extrema of the loss on the ground truth, is also a local extremum of the Laplacian of the loss. As it is evident \textcolor{black}{from the bottom panels of} Fig.~\ref{ResUNetAAllLossnLaplacians} this is not the case for all loss functions.

\textcolor{black}{In more detail, in Fig.~\ref{ResUNetAAllLossnLaplacians}, we plot the gradient field of the  Dice loss functions and the  corresponding Laplacian fields. In the top row are shown the gradient fields of the  three different functional form of the Dice loss and the form with their complements. From left to right we have the   Dice coefficient based loss with summation of probabilities in the denumerator, $D_1(p,l)$,  its complement, $\tilde{D}_1(p,l)$, the  Dice loss with summation of squares in the denominator, $D_2(p,l)$, its complement, $\tilde{D}_2(p,l)$, and the third form of the Dice loss with summation of squares that also includes a subtraction term,  $D_3(p,l)$, and its complement, $\tilde{D}_3(p,l)$.  
 From the gradient flow of the $D1$ loss, it is evident that for a random initialization of the network weights (which is the case in deep learning) that corresponds to some random point $(p_x,p_y)$ of the loss landscape, the gradients of the loss with respect to $p_x, p_y$ will not necessarily direct to the ground truth point in $(1,0)$. In this respect, the generalized Dice loss with complement, $\tilde{D}_1$  behaves better. However,  the gradient flow lines do not pass through  the ground truth point for all possible pairs of values $(p_x,p_y)$. 
For the case of the loss functions $D_2$, $D_3$ and their complements, the gradient flow lines pass through the ground truth point, but these are not straight lines. Their forms with complement, $\tilde{D}_2,\tilde{D}_3$, have gradient lines flowing straight towards the ground truth irrespective of the (random) initialization point. 
The Laplacians of these loss functions are in the corresponding bottom panels of Fig.~\ref{ResUNetAAllLossnLaplacians}. 
It is clear that the extremum of the Laplacian operator is closer to the ground truth values only for the cases where we consider the loss functions with complement.  Interestingly, the Laplacian of the Tanimoto functional form ($D_3$) has extremum values closer to the ground truth point in comparison with the $D_2$ functional form.}

\textcolor{black}{In summary,  the Tanimoto loss with complement has  gradient flow lines that are straight lines (geodesics, i.e. they follow the shortest path) pointing to the ground truth from any random initialization point, and the second order derivative has extremum on the location of the ground truth.
This demonstrates, according to our opinion, the superiority of the Tanimoto with complement as a loss function, among the family of loss functions based on the Dice coefficient, for training deep learning models.}

\begin{figure}
\centering
\includegraphics[width=\columnwidth]{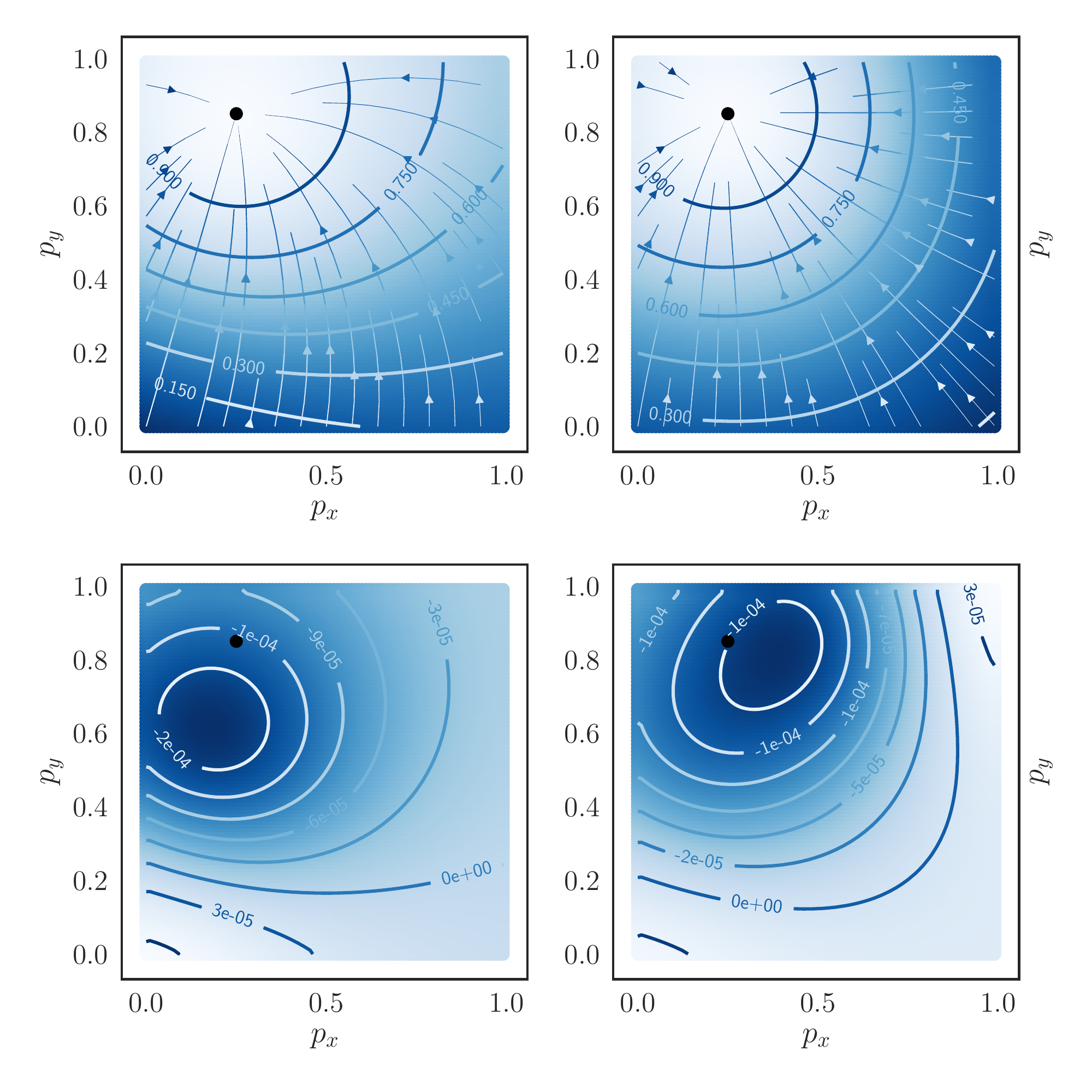}
\caption{\textcolor{black}{Top panels: gradient flow of the Tanimoto (top left) and Tanimoto with complement (top right) loss functions for a continuous target value. Bottom panels: corresponding Laplacian of the gradients. The ``ground truth'', $(0.25,0.85)$ is represented with a black dot. \textcolor{black}{The numerical values of the isocontours on the images describe numerically the colorscheme with darker values corresponding to lower values.}}}
\label{Tnmt_wth_Dual_contLoss}
\end{figure}

\subsubsection{Tanimoto with complement as a regression loss}

It should be stressed, that if we restrict the output of the neural network in the range $[0,1]$ (with the use of softmax or sigmoid activations) then the Tanimoto loss can be used to recover also continuous variables in the range $[0,1]$. In Fig.~\ref{Tnmt_wth_Dual_contLoss} we present an example of this, for a ground truth vector of $l=(0.25, 0.85)$. In the top panels, we plot the gradient flow of the Tanimoto (left) and Tanimoto with complement (right) functions. In the bottom panels,  we plot the corresponding functions obtained after applying the Laplacian operator to the loss functions. This is an appealing property for the case of multi-task learning, where one of the complementary goals is a continuous loss function.  The reason being that the gradients of these components will have  similar magnitude scale and the training will be equally balanced  to all complementary tasks. In contrast, when we use different functional form functions for different tasks in the optimization, we have to explicitely balance the gradients of the different components, with the extra cost of having to find the additional hyperparameter(s). For example, assuming we have two complementary tasks described by two different functional form functions, $L_1$ and  $L_2$, then the total loss must be balanced with the usage of some (unknown) hyperparameter $a$ that needs to be calculated: $L_{\mathrm{total}} = L_1 + a L_2$. 

\begin{figure}
\centering
\includegraphics[width=\columnwidth,trim=.5cm 0.5cm 0.5cm 0.5cm,clip]{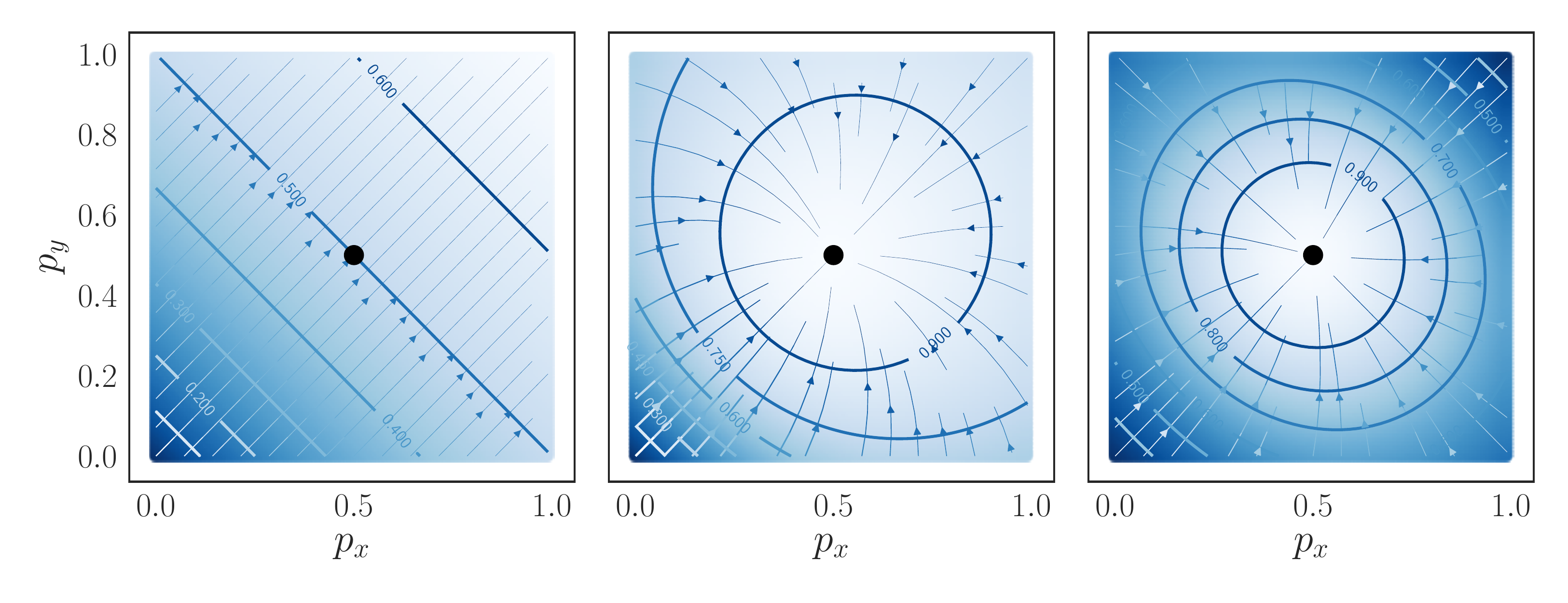}
\caption{\textcolor{black}{Gradient flow of the Dice family of losses for three different functional forms.  From left to right: standard Dice loss ($D_1$, Eq. \eqref{eqn:dice1}, \citealt{DBLP:journals/corr/SudreLVOC17},  Dice loss with squares in the denominator ($D_2$, Eq. \eqref{eqn:dice2},  \citealt{DBLP:journals/corr/MilletariNA16}), Tanimoto with complement (Eq, \eqref{Tanimoto_wth_dual}, this work). The ``ground truth'', $(0.5,0.5)$ is represented with a black dot. It is clear that Tanimoto with complement has gradient flow (i.e. gradient magnitudes and direction) that is symmetric around the ground truth point, thus making it suitable for continuous regression problems. In contrast the Dice loss $D_1$ is not suitable for this usage, while $D_2$ has a clear assymetry that affects the gradients magnitude around the ground truth.}}
\label{Dice1_vs_Dice2_vs_TwD}
\end{figure}

In Fig. \ref{Dice1_vs_Dice2_vs_TwD} we plot the gradient flow for three different members of the Dice family loss functional forms. From left to right we plot  the standard Dice loss with summation in the denominator ($D_1$, Eq. \eqref{eqn:dice1}, \citealt{DBLP:journals/corr/SudreLVOC17}), the  Dice loss with squares in the denominator ($D_2$, Eq. \eqref{eqn:dice2}, \citealt{DBLP:journals/corr/MilletariNA16}) and  Tanimoto with complement (Eq, \eqref{Tanimoto_wth_dual} that we introduce in this work. It is clear that  the Tanimoto with complement has the highest degree of symmetric gradients in both magnitude and direction around the ground truth point (for this example, the ``ground truth'' is $(p_x, p_y)=(0.5, 0.5)$). In addition, it also has steeper gradients as this is demonstrated from the distance of isocontours. The above help achieving faster convergence in problems with gradient descent optimization.

\subsubsection{Generalization to multiclass imbalanced problems}
Following the Dice loss modification of \cite{DBLP:journals/corr/SudreLVOC17} for including weights per class, we generalize the Tanimoto loss for multi-class labelling of images: 
\begin{equation}
\label{tanimoto_loss_simple}
T(p_{iJ},l_{iJ}) = 
\frac{\sum_{J=1}^{N_{\mathrm{class}}} w_J \sum_{i=1}^{N_{\mathrm{pixels}}} p_{iJ} l_{iJ}}{ 
\sum_{J=1}^{N_{\mathrm{class}}} w_J 
\sum_{i=1}^{N_{\mathrm{pixels}}} \left( p_{iJ}^2 +  l_{iJ}^2 -  p_{iJ} l_{iJ} \right) }.
\end{equation}
Here $w_J$ are the weights per class $J$, $p_{iJ}$ is the probability of pixel $i$ belonging to class $J$ and $l_{iJ}$ is the label of pixel $i$ belonging to class $J$. Weights are derived following the inverse  ``volume'' weighting scheme per \citet{journals/tmi/CrumCH06}:
\begin{equation}
w_J = V_J^{-2},
\end{equation}
where $V_J $ is the total sum of true positives per class $J$, $V_J = \sum_{i=1}^{N_{\mathrm{pixels}}} l_{iJ}$. 
\textcolor{black}{
In the following we will exclusively use the weighted Tanimoto (Eq. \ref{tanimoto_loss_simple}) with complement, $\tilde{T}(p_{iJ},l_{iJ}) = (T(p_{iJ},l_{iJ}) + T(\mathbf{1}-p_{iJ},\mathbf{1}-l_{iJ}))/2$, and we will refer to it simply as the Tanimoto loss.  }
  
\subsection{Data augmentation}
\label{resuneta_data_augmentation}


To avoid overfitting, we relied on geometric data augmentation so that, in each iteration, the algorithm never sees the exact same set of images (i.e. the batch of images is always different). Each pair of image and ground truth mask are rotated at a random angle, with a random centre and zoomed in/out according to a random scale factor. The parts of the image that are left out from the frame after the transformation are filled in with reflect padding. \textcolor{black}{This data augmentation methodology is particularly useful for aerial images of urban areas due to the high degree of reflect symmetry these areas have by design.}
We also used random reflections in $x$, $y$ directions as an additional data augmentation routine. 

\begin{figure}[h!]
\centering
\includegraphics[width=\linewidth]{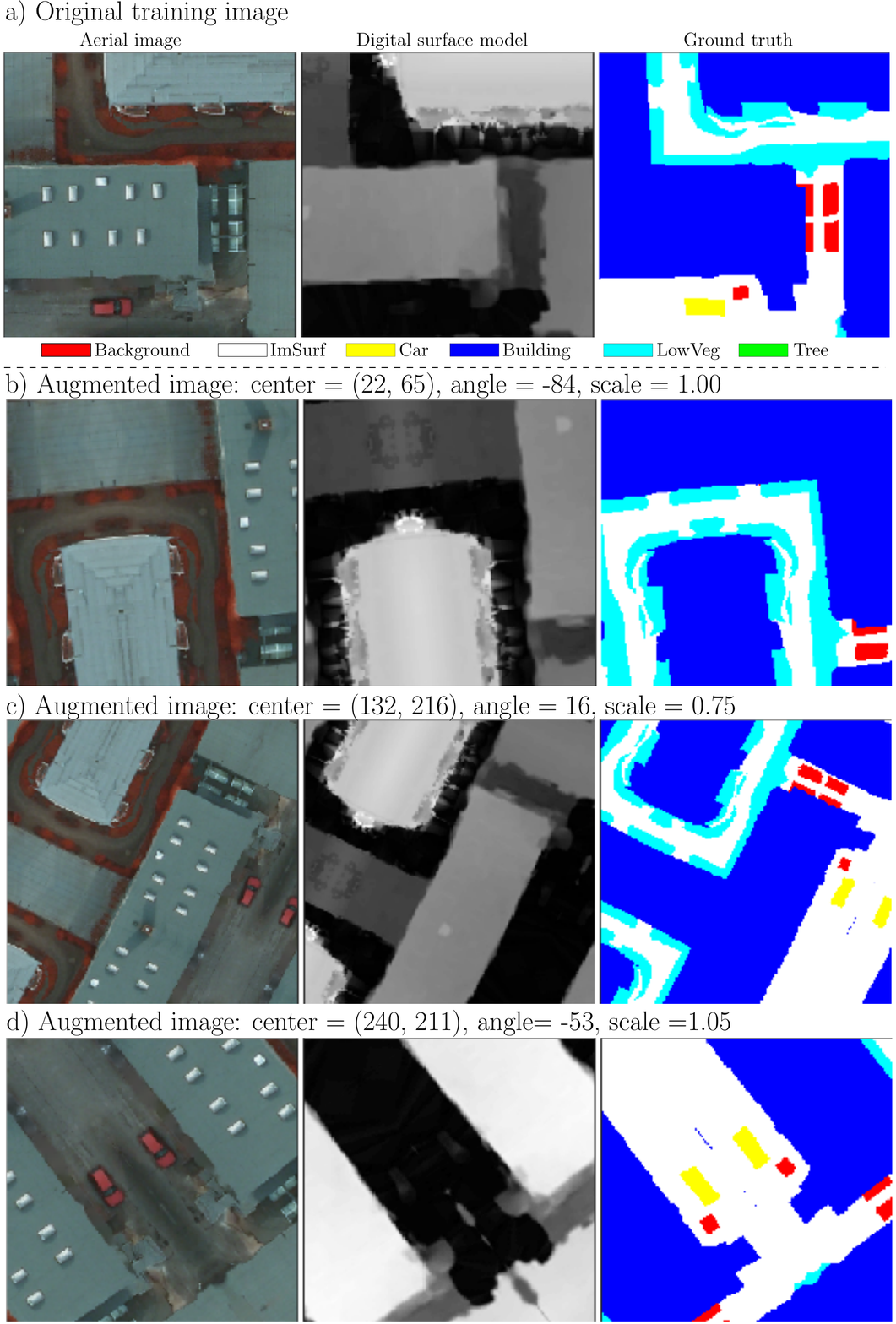}
\caption{Example of data augmentation on image patches of size 256x256 \textcolor{black}{(ground sampling distance 10cm - FoV$\times$4 dataset)}. Top row: original image, subsequent rows: random rotations with respect to (random) center and at a random scale (zoom in/out). Reflect padding was used to fill the missing values of the image after the transformation.}
\label{resuneta_potsdam_random_trans}
\end{figure}

The regularization approach is illustrated in Fig. \ref{resuneta_potsdam_random_trans} for a single datum of the ISPRS Potsdam data set (top row, \textcolor{black}{FoV$\times$4 dataset}). From left to right, we show the false color infrared image of a 256x256 image patch, the corresponding digital elevation model, and the ground truth mask. In rows 2-4, we provide examples of the random transformations of the original image.  By exposing the algorithm to different perspectives of the same objects scenery, we encode the prior knowledge that the algorithm should be able to identify the objects for all possible affine transformations. That is, we make the segmentation task invariant in affine transformations. This is quite similar to the functionality of the Spatial Transformer Network \citep{DBLP:journals/corr/JaderbergSZK15}, with the difference that this information is hard-coded in the data rather than the internal layers of the network. It should be noted that several authors report performance gains when they use inputs viewed at different scales, \eg \cite{DBLP:journals/corr/AudebertSL16a} and \cite{rs10111768}.

\subsection{Inference methodology}
\label{resuneta_inference}

In this section, we detail the approach we followed for performing inference  over large true orthophoto that exceeds the 256x256 size of the image patches we use during training.

As detailed in the introduction, FCNs such as \resuneta use contextual information to increase their performance. In practice, this means that in a single 256x256 window for inference, the pixels that are closer to the edges will not be classified as confidently as the ones close to the center because more contextual information is available to central pixels. Indeed,  contextual information for the edge pixels is limited since there is no information outside the boundaries of the image patch. To further improve the performance of the algorithm and provide seamless segmentation masks, the inference is enhanced with multiple overlapping inference windows.  This is like deciding on the  classification result from multiple views (sliding windows) of the same objects. This type of approach is also used for large-scale land cover classification to combine classification in a seamless map~\citep{lambert2016cropland, waldner2017national}.

Practically, we perform multiple overlapping windows passes over the whole tile and store the class probabilities for each pixel and each pass. The final class probability vector ($\tilde{p}_i(x,y)$) is obtained using the average of all the prediction views.
The sliding window has size equal to the tile dimensions (256x256), however, we step through the whole image in strides of  256/4 = 64 pixels, in order to get multiple inference probabilities for each pixel. In order to account for the lack of information outside the tile boundaries, we pad each tile with reflect padding at a size equal to 256/2 = 128 pixels \citep{DBLP:journals/corr/RonnebergerFB15}.

\section{Data and preprocessing}
\label{resuneta_data}

We sourced data from the ISPRS 2D Semantic Labelling Challenge and in particular the Potsdam data set~\citep{ISPRS}. The data consist of a set of true orthophoto (TOP) extracted from a larger mosaic, and a Digital Surface Model (DSM). The TOP consists of the four spectral bands in the visible (VIS; red (R), green (G), and blue (G) and in the near infrared (NIR) and the ground sampling distance is 5 cm. The normalized DSM layer provides information on the height of each pixel as the ground elevation was subtracted. The four spectral bands (VISNIR) and the normalized DSM were stacked (VISNIR+DSM) to be used to train the semantic segmentation models. The labels consist of six classes, namely impervious surfaces, buildings, cars, low vegetation, trees, and background.

Unlike conventional pixel-based (e.g. random forests) or GEOBIA approaches, CNNs have the ability to ``see'' image objects in their contexts, which provides additional information to discriminate between classes. Thus, working with large image patches maximizes the competitive advantage of CNNs, however,  limits to the maximum patch size are dictated by memory restrictions of the GPU hardware.  
We have created two versions of the training data. In the first version,  we resampled the image tiles to half their original resolution and extracted image patches of size 256x256 pixels to train the network. The reduction of the original tile size to half was decided with the mindset that we can include more context information per image patch. This resulted in image patches with four times larger Field of View (hereafter FoV) for the same 256x256 patch size. \textcolor{black}{We will refer to this dataset as (FoV$\times$4) as it includes 4 times larger Field of View (area) in a single 256x256 image patch (in comparison with 256x256 image patches extracted directly from the original unscaled dataset). In the second version of the training data, we kept the full resolution tiles and again extracted image patches of size 256x256 pixels. We will refer to this dataset as FoV$\times$1.}  
The 256x256 image patch size was the maximum size that the memory capacity of our hardware configuration could handle (see \ref{resuneta_babysitting}) so as to process a meaningfully large batch of datums. Each of the 256x256  patches used for training was extracted from a sliding window swiped over the whole tile at a stride, \ie step, of 128 pixels. This approach guarantees that all pixels at the edge of a patch become central pixels in subsequent patches. After slicing the original images, we split\footnote{Making sure there is no overlap between the image patches of the training, validation and test sets.} the 256x256 patches into a training set, a validation set, and a test set with the following ratios: 0.8-0.1-0.1. 

\textcolor{black}{
The purpose of the two distinct datasets is: the FoV$\times$4 is useful in order to understand how much (if any) the increased context information improves the performance of the algorithm. It also allows us to perform more experiments much faster due to the decreased volume of data. The FoV$\times$4 dataset is approximately $\sim$50GB, with $\sim$10k of pairs of images, masks. The FoV$\times$1 has volume size of $\sim$250GB, and $\sim$40k pairs of images, masks.  
In addition, the FoV$\times$4  is a useful benchmark on how the algorithm behaves with a smaller amount of data than the one provided.  
Finally, the FoV$\times$1 version is used in order to compare the performance of our architecture with other published results.}

\section{Architecture and Tanimoto loss experimental analysis}
\label{ablation_study_all}
In this section, we perform an experimental analysis of the \resuneta architecture as well as the performance of the Tanimoto with complement loss function. 

\subsection{\textcolor{black}{Accuracy assessment}}

For each tile of the test set, we constructed the confusion matrix and extracted the several accuracy metrics such as the overall accuracy (OA), the precision, the recall, and the F1-score ($F_1$):
\begin{align}
\mathrm{OA} &= \frac{TP+TN}{FP+FN}\\
\mathrm{precision} &= \frac{TP}{TP+FP}\\
\mathrm{recall} &= \frac{TP}{TP+FN} \\
F_1 &= 2 \cdot \frac{\mathrm{precision} \cdot \mathrm{recall}}{\mathrm{precision}+\mathrm{recall}}
\end{align}
where $TP$,  $FP$, $FN$, and  $TN$ are the is true positive, false positive, false negative and true negative classifications, respectively.

In addition, for the validation dataset (for which we have ground truth labels), we use the Matthews Correlation Coefficient (hereafter MCC, \citealt{MATTHEWS1975442}):
\begin{equation}
\label{mcc_def}
\mathrm{MCC}=\frac{TP \times TN - FP \times FN}{\sqrt{(TP+FP) (TP+FN) (TN+FP)(TN+FN) }}
\end{equation}

\subsection{Architecture ablation study}
\label{arch_ablation_study}
In this section, we design two experiments in order to evaluate the performance of the various modules that we use in the \texttt{ResUNet-a} architecture. \textcolor{black}{In these experiments we used the FoV$\times$1 dataset, as this is the dataset that will be the ultimate testbed of \resuneta performance against other modeling fra-meworks. Our metric for understanding the performance gains of the various models tested is the model complexity and training convergence: if model A has greater (or equal) number of parameters than model B, and model A converges faster to optimality than model B, then it is most likely that will also achieve the highest overall score.}

\begin{figure}
\centering
\includegraphics[width=\columnwidth]{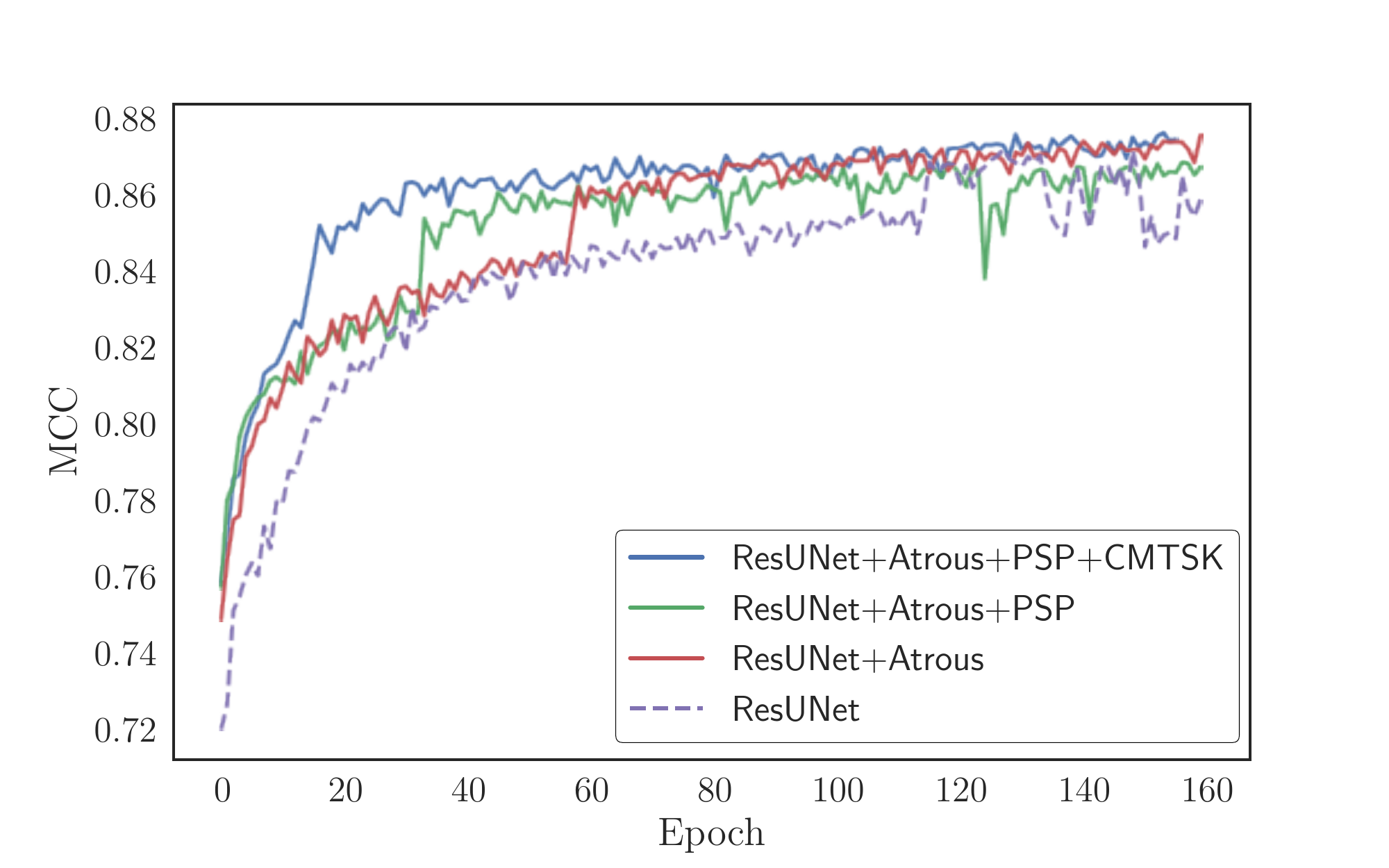}
\caption{Convergence performance of the \resuneta architecture. Starting from a baseline wide-ResUNet, we add components keeping all training hyper-parameters identical.  }
\label{Ablation_fig1}
\end{figure}

\textcolor{black}{In the first experiment we test the convergent properties of our architecture. In this, we are not interested in the final performance (after learning rate reduction and finetuning) which is a very time consuming operation, but how \resuneta behaves during training for the same fixed set of hyperparameters and epochs.} 
We start by training a baseline model, a modified \texttt{ResUNet} \citep{DBLP:journals/corr/abs-1711-10684}  where in order to keep the number of parameters identical with the case with atrous, we use the same \texttt{ResUNet-a} building blocks with dilation rate equal to 1 for all parallel residual blocks (i.e. there are no atrous convolutions). This is similar in philosophy with the wide residual networks \citep{DBLP:journals/corr/ZagoruykoK16}, however, there are no dropout layers. Then, we modify this baseline by increasing the dilation rate, thus adding atrous convolutions (model: \texttt{ResUNet} + \texttt{Atrous}). It should be clear that the only difference between the models \texttt{ResUNet} and \texttt{ResUNet + Atrous} is that the latter has different dilation rates than the former, i.e. they have identical number of parameters. 
Then we add PSPPooling in both the middle and the end of the framework (model: \texttt{ResUNet + Atrous + PSP}), and finally we apply the conditioned multitasking, i.e. the full \resuneta model (model: \texttt{ResUNet} + \texttt{Atrous} + \texttt{PSP} + \texttt{CMTSK}). The differences in performance of the convergence rates is incremental with each module addition. This  performance difference can be seen in Fig. \ref{Ablation_fig1} and is substantial. In Fig. \ref{Ablation_fig1}  
we plot the Matthews correlation coefficient (MCC) for all models. The MCC was calculated using the success rate over all classes. The baseline \texttt{ResUNet} requires approximately 120 epochs to achieve the same performance level that \texttt{ResUNet-a - cmtsk} achieves in epoch $\sim 40$. The mere change from simple (\texttt{ResUNet}) to atrous  convolutions (model \texttt{ResUNet}+ Atrous) almost doubles the convergence rate. The inclusion of the PSP module (both middle and end) provides additional learning capacity, however,  it also comes with training instability. This is fixed by adding the conditioned multitasking module in the final model.   \textcolor{black}{Clearly, each module addition: (a) increases the complexity of the model since it increases the total number of parameters and (b) it improves the convergence performance.}

\begin{figure}
\centering
\includegraphics[width=\columnwidth]{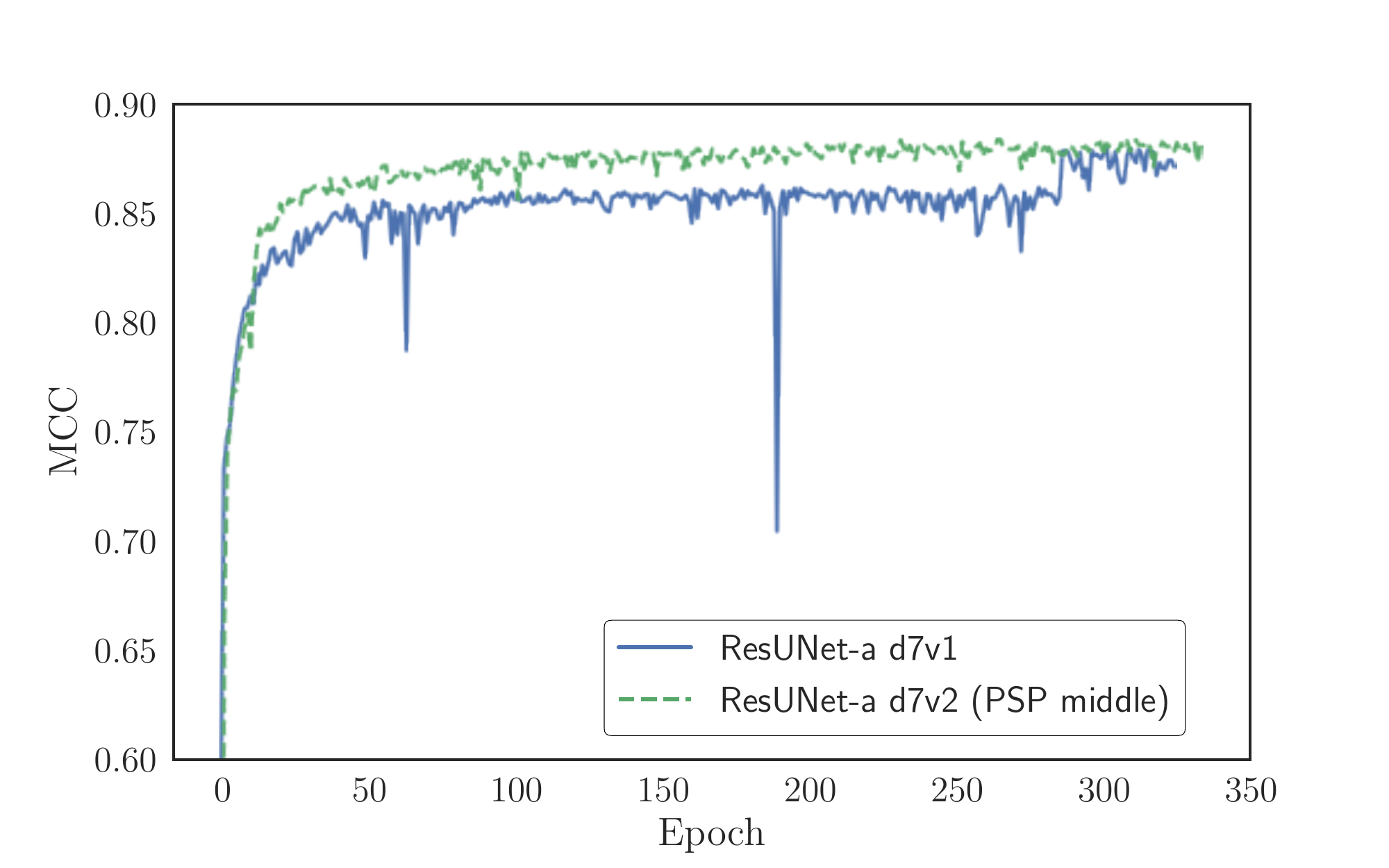}
\caption{Convergence performance evaluation for the PSPPooling layer in the middle. We compare two architectures \resuneta d7v1 architecture without PSPPooling layer (blue solid line) and with PSPPooling layer in the middle (i.e. \textcolor{black}{d7v2}, dashed green line).  It is clear that the insertion of the PSPPooling layer  in the middle of the architecture boosts convergence performance of the network.}
\label{Ablation_fig2}
\end{figure}

Next, we are interested in evaluating the importance of the PSPPooling layer. In our experiments we found this layer to be more important in the middle of the network than before the last output layers. For this purpose, we train two \resuneta d7 models, the d7v1 and d7v2, that are identical in all aspects except that the latter has a PSPPooling layer in the middle. Both models are trained with the same fixed set of hyperparameters (i.e. no learning rate reduction takes place during training).  In Fig \ref{Ablation_fig2} we show the convergence evolution of these networks. It is clear that the model with the PSPPooling layer in the middle (i.e. v2) converges much faster to optimality, despite the fact that it has greater complexity (i.e. number of parameters) than the model d7v1.  

\begin{figure}
\centering
\includegraphics[width=\columnwidth,trim=1.5cm 1.5cm 1.5cm 1.5cm,clip]{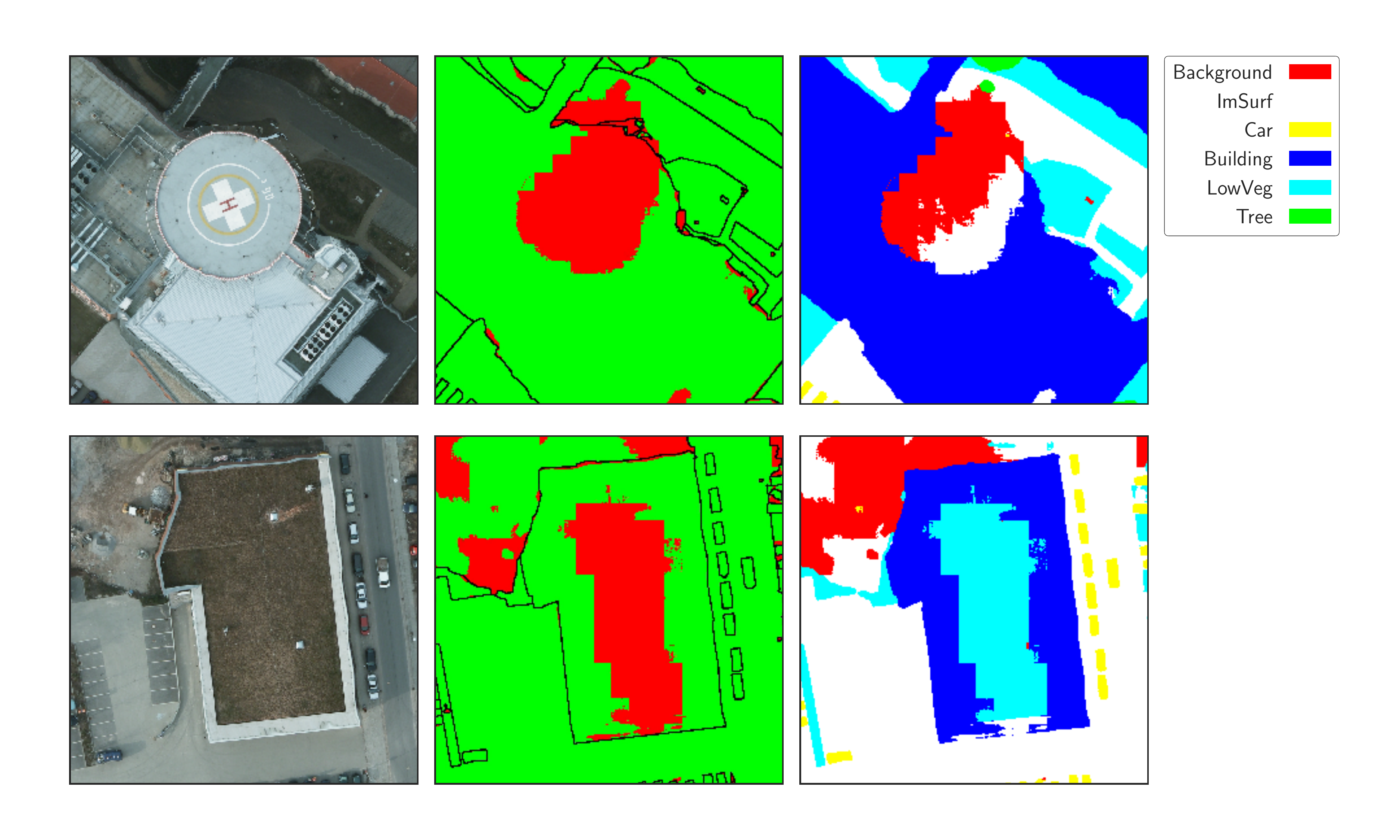}
\caption{\textcolor{black}{Example of PSPPooling erroneous inference behaviour for segmentation tasks. The error appears in the form of (parts of) squares blocks. For each row, from left to right: RGB bands of input image, error map, and segmentation mask. The top row corresponds to a zoom in region of tile 6\_14 and the bottom to a region of tile 6\_15. Each image patch is of size 1256$\times$1256 and corresponds to a ground sampling distance of 5cm.}}
\label{PSPerrorInferenceExample}
\end{figure}

 In addition to the above, we have to note that when the network performs erroneous inference, due to the PSPPooling layer in the end, this may appear in the form of square blocks, indicating the influence of the pooling area in square subregions of the output. The last PSPPooling layer is in particular problematic when dealing with regression output problems. 
 This is the reason why we did not use it in the evaluation of color and distance transform modules in the multitasking networks. 
\textcolor{black}{In Fig.~\ref{PSPerrorInferenceExample} we present two examples of erroneous inference of the last PSPPooling layer that appear in the form of square blocks. The first row corresponds to a zoom in region of tile 6\_14, and the bottom row to a zoom in region of tile 6\_15. From left to right: RGB bands of input image, error map, and inference map. From the boundary of the error map It can be seen that the boundary of the error map has areas that appear in the form of square blocks. That is, the effect of the pooling operation in various scales can dominate the inference area.}

\begin{figure}
\centering
\includegraphics[width=\columnwidth]{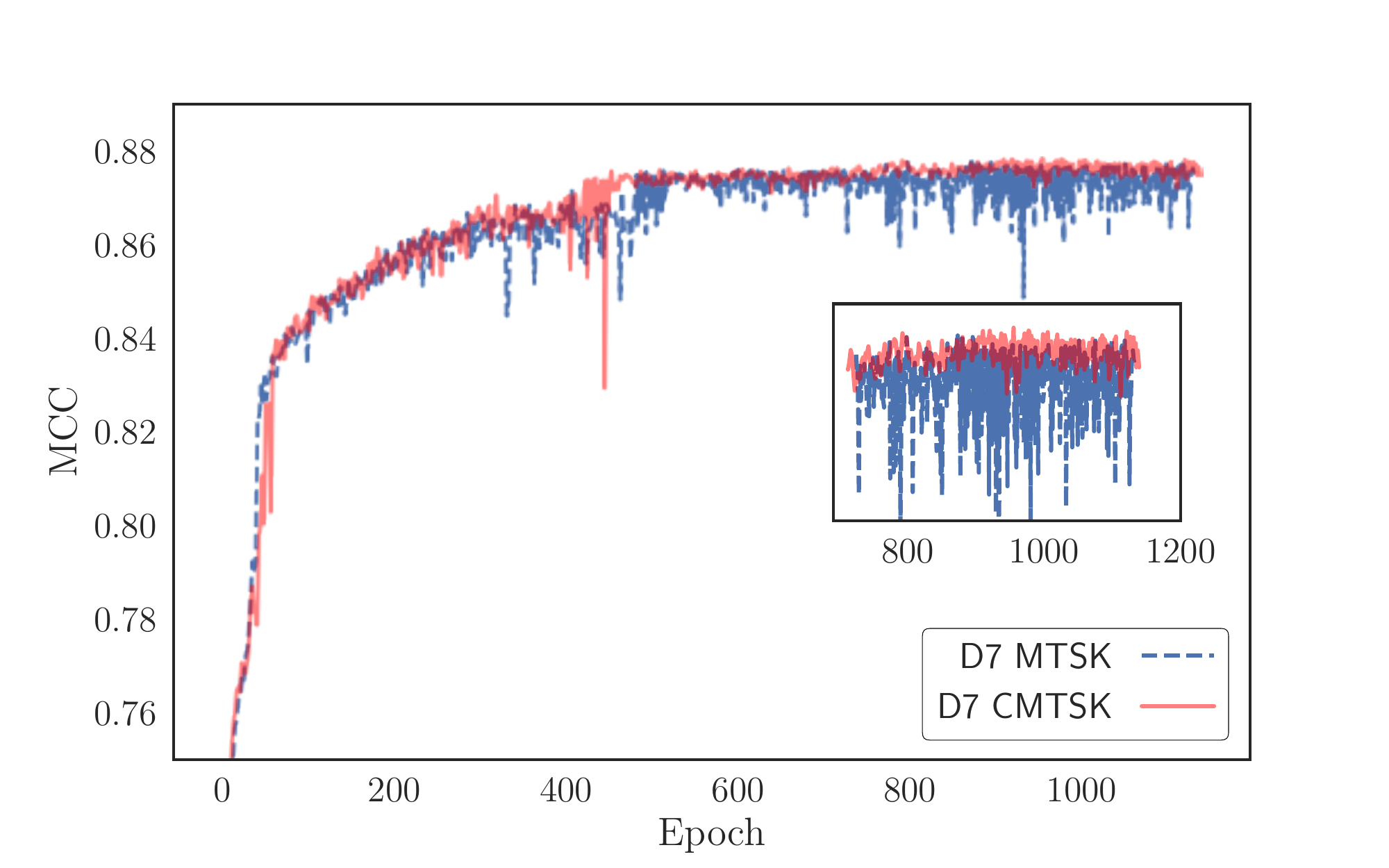}
\caption{Training evolution of the \textcolor{black}{conditioned} vs the standard multi-task models for the \resuneta d7\textcolor{black}{v1} family of models. The \textcolor{black}{conditioned}  model (cmtsk) is represented with a red solid line, while the standard multi-task (mtsk) one with a dashed blue line. The mtsk demonstrates higher variance during training especially closer to the final convergence.}
\label{MTSK_vs_CMTSK}
\end{figure}

Comparing \resunetamtsk and \resunetacmtsk models \textcolor{black}{(on the basis of the d7v1 feature extractor)}, we find that the latter demonstrates smaller variance in the values of the loss function (and in consequence, the performance metric) during training. In Fig. \ref{MTSK_vs_CMTSK} we present an example of the comparative training evolution of the \resuneta d7\textcolor{black}{v1} mtsk versus the \resuneta d7\textcolor{black}{v1} cmtsk models.  
\textcolor{black}{It is clear that the conditioned inference model demonstrates smaller variance, and that, despite the random fluctuations of the MCC coefficient, the median performance of the conditioned multitasking model is higher than the median performance of the simple multitasking model.}
This helps in stabilizing the gradient updates and results slightly better performance. 
We have also found that the inclusion of the identity reconstruction of the input image (in HSV colorspace) helps further to reduce the variance of the performance metric.  

Our conclusion is that the greatest gain in using the  \textcolor{black}{conditioned} multi-task model is in faster and consistent convergence to optimal values, as well as better segmentation of the boundaries (in comparison with  the single output models).

\subsection{Performance evaluation of the proposed loss function}

\begin{figure}
\centering
\includegraphics[width=\columnwidth]{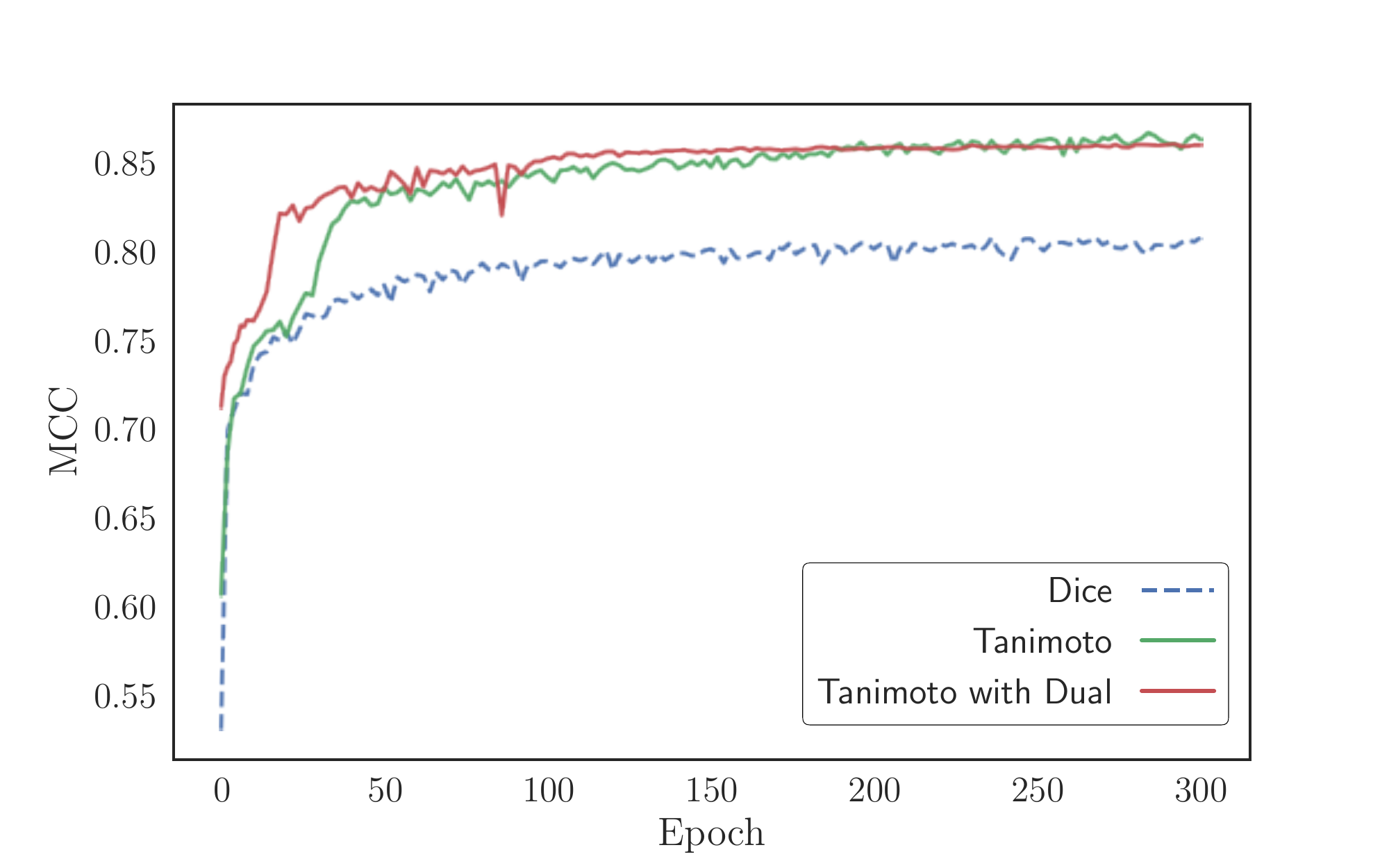}
\caption{\textcolor{black}{Training evolution of the same model using three different loss functions. The Tanimoto with complement loss (solid red line, this work), the Tanimoto (solid green line), and the Dice loss} 
\citep[dashed blue line,][]{DBLP:journals/corr/SudreLVOC17}.}
\label{Dice_vs_TnmtD}
\end{figure}

In order to demonstrate the performance difference between the Dice loss as defined in \cite{journals/tmi/CrumCH06} and the Tanimoto loss, and the Tanimoto with complement (Eq. \ref{Tanimoto_wth_dual}) we train three identical models with the same set of hyper-parameters. The weighting scheme is the same for all losses. \textcolor{black}{In this experiment we used the FoV$\times$4 dataset, in order to complete it shorter time. As the loss function cannot be responsible for overfitting (only the model capacity can lead to such behavior) our results persist also with the larger FoV$\times$1 dataset}.  
In Fig. \ref{Dice_vs_TnmtD} we plot the Matthews correlation coefficient (MCC).  In this particular example, we are not interested in achieving maximum performance  by reducing the learning rate and pushing the boundaries of what the model can achieve. We are only interested to compare the relative performance for the same training epochs between the different losses with an identical set of  fixed hyperparameters. 
It is evident that the Dice loss stagnates to lower values, while the Tanimoto loss with complement converges faster to an optimal value.   The difference in performance is significant: the Tanimoto loss with complement achieves for the same number of epochs an  $\mathrm{MCC} = 85.99$, while the Dice loss stagnates at $\mathrm{MCC}= 80.72$. The Tanimoto loss without complement (Eq. \ref{tanimoto_loss_simple}) gives a similar performance with the Tanimoto with complement, however, it converges relatively slower and demonstrates greater variance. In all experiments we performed, the Tanimoto with complement gave us the best performance.

\begin{table*}
\footnotesize
\caption{\textcolor{black}{Potsdam comparison of results (F1 score and overall accuracy - OA) for the various \resuneta models trained on the FoV$\times$4 and FoV$\times$1 datasets. Highest score is marked with bold.  The average F1-score was calculated using all classes except the ``Background'' class. The overall accuracy, was calculated including the ``Background'' category.}}
\label{resuneta_between_comparison}
\begin{center}
\begin{tabular}{ l c c  c  c  c  c  c  c  }\hline
Methods  &DataSet	&ImSurface	&Building &LowVeg &Tree &Car &Avg. F1 & OA\\ \hline 
\resuneta d6 &(FoV$\times$4) & 92.7 & 97.1 & 86.4 & 85.8 &  95.8 & 91.6 & 90.1 \\
\resuneta d6 cmtsk &(FoV$\times$4) & 91.4 & {\bf 97.6} & 87.4 & 88.1 &  95.3 & 91.9 & 90.1 \\[2pt] \hline
\resuneta d7v1 mtsk &(FoV$\times$4) & 92.9 & 97.2 & 86.8 & 87.4 &  96.0 & 92.1 & 90.6 \\
\resuneta d7v1 cmtsk &(FoV$\times$4)  & 92.9 & 97.2 & 87.0 & 87.5 &  95.8 & 92.1 & 90.7 \\[2pt]
\hline \hline
\resuneta d6 cmtsk &(FoV$\times$1) & 93.0 & 97.2 & 87.5 & 88.4 & 96.1 & 92.4 & 91.0\\ 
\resuneta d7v2 cmtsk &(FoV$\times$1) & \textbf{93.5} & 97.2 & \textbf{88.2} & \textbf{89.2} & \textbf{96.4} & \textbf{92.9} & \textbf{91.5}\\\hline 
\end{tabular}
\end{center}
\end{table*}

\begin{figure*}[h!]
\centering
\includegraphics[width=\textwidth,trim=1.5cm 5.5cm 1.125cm 2.0125cm,clip]{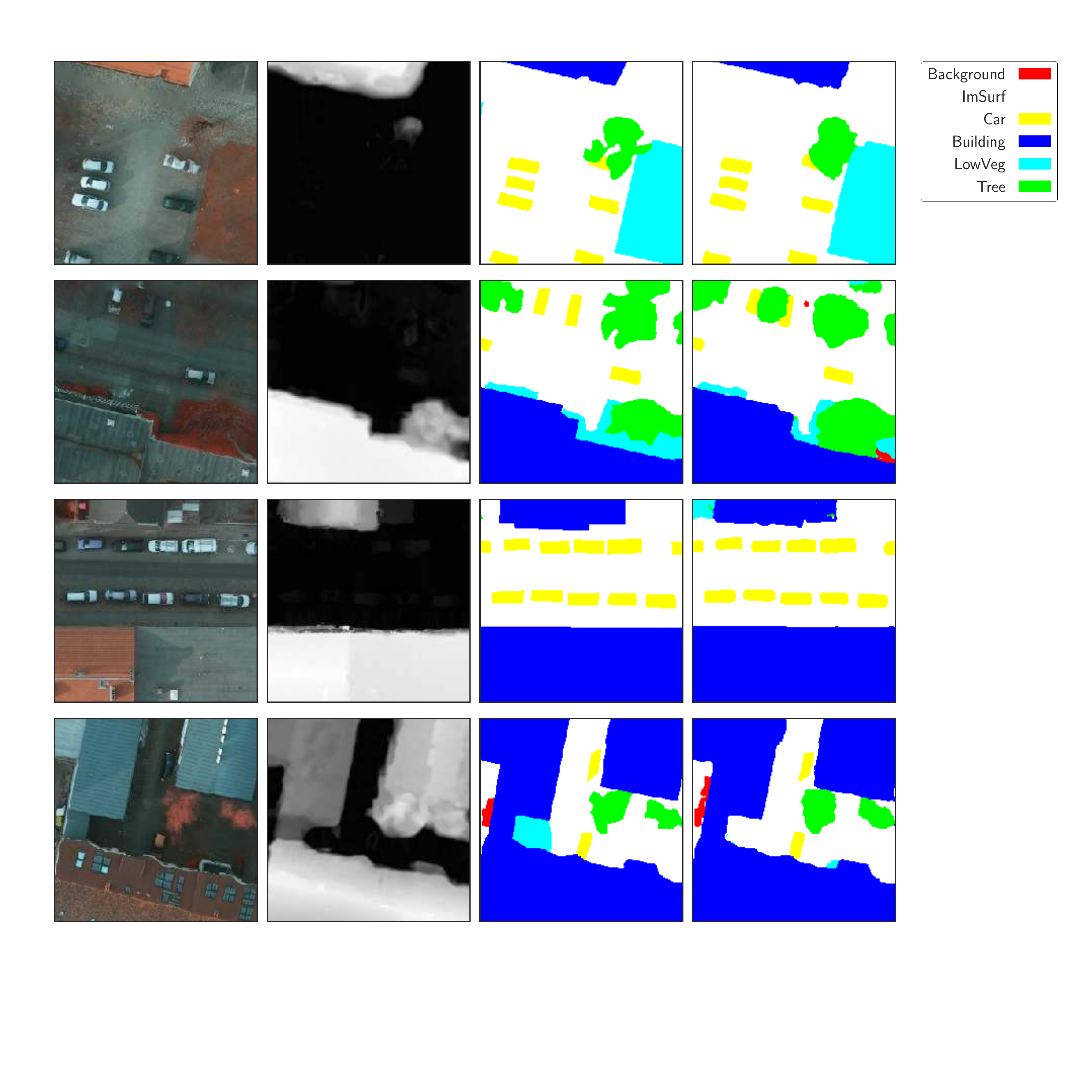}

\caption{\resuneta d7v1 cmtsk inference on unseen test patches of size 256x256 \textcolor{black}{(FoV$\times$4 - \textcolor{black}{ground sampling distance 10cm} )}. From left to right: rgb image, digital elevation map, ground truth, and prediction.}
\label{resuneta_potsdam_2_13}
\end{figure*}

\begin{figure}[h!]
\centering
\includegraphics[width=\columnwidth]{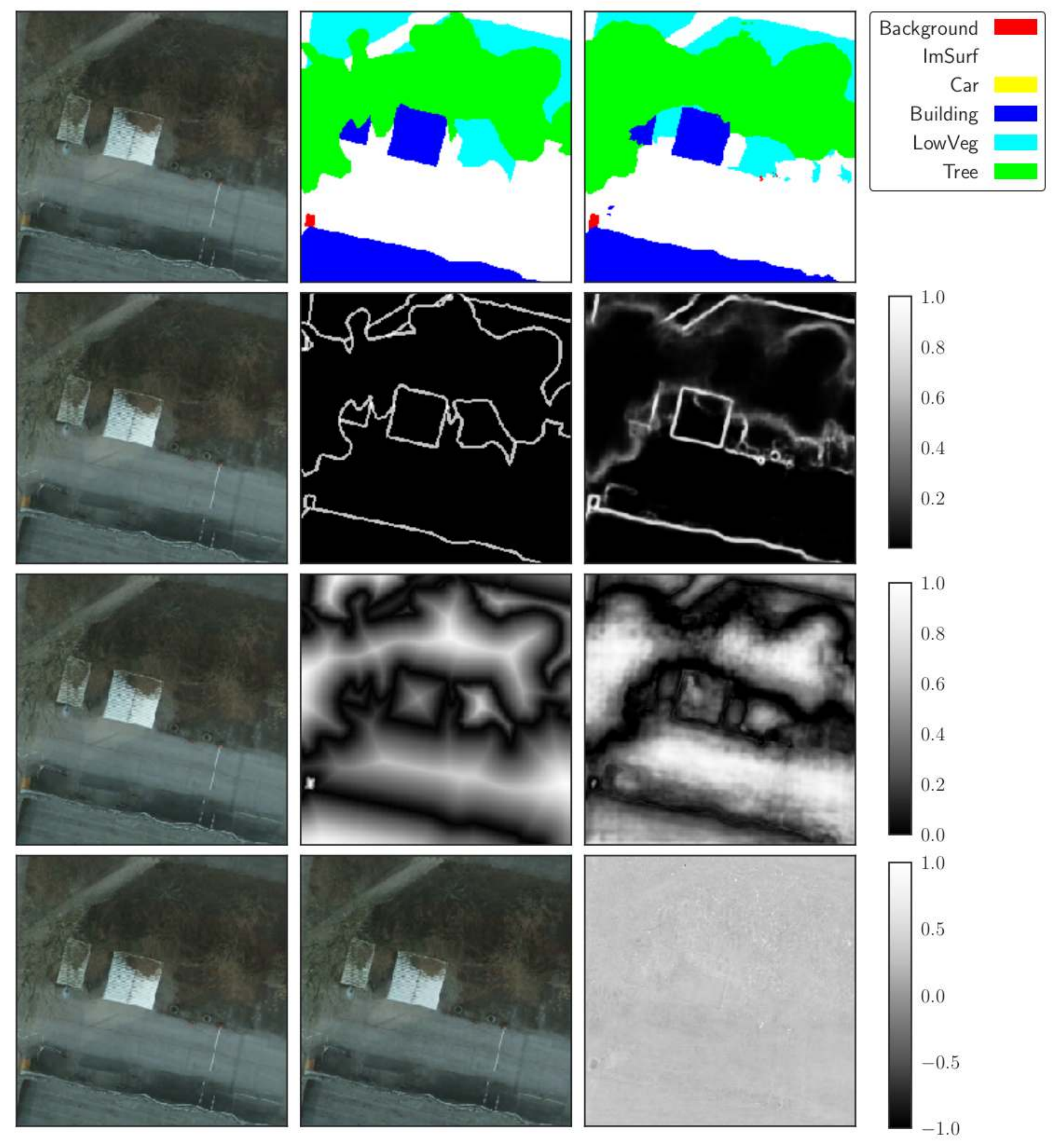}
\caption{\resuneta d7\textcolor{black}{v1} cmtsk all tasks inference on unseen test patches of size 256x256 for the \textcolor{black}{FoV$\times$4 dataset \textcolor{black}{(ground sampling distance 10cm)}. From left to right, top row: input image, ground truth segmentation mask, predicted segmentation mask. Second row: input image, ground truth boundaries, predicted boundaries (confidence). Third row: input image, ground truth distance map, inferred distance map. Bottom row: input image, reconstructed image, difference between input and predicted image. }}
\label{resuneta_inference_all_demo}
\end{figure}

\section{Results and discussion}
\label{resuneta_results}

In this section, we present and discuss  the performance of \texttt{ResUNet-a}. We also compare the efficiency of our model with results from architectures of other authors. 
We present results in both the FoV$\times$4 and FoV$\times$1 versions of the ISPRS Potsdam dataset. 
It should be noted that the ground truth masks of the test set were made publicly available on June 2018.  Since then, the \cite{ISPRS} 2D semantic label online test results are  not being updated. The ground truth labels used to calculate the performance scores are the ones with the eroded boundaries.

\subsection{Design of experiments}

\textcolor{black}{
In Section \ref{ablation_study_all} we tested the convergence properties of the various modules that \resuneta uses. In this section, our goal is to train the best convergent models until optimality and compare their performance. To this aim we document the following set of representative experiments: (a) \resuneta d6 vs \resuneta d6 cmtsk. The relative comparison of this will give us the performance boost between single task and conditioned multitasking models, keeping everything else the same. (b) \resuneta d7v1 mtsk vs \resuneta d7v1 conditioned mtsk. Here we are trying to see if conditioned multitasking improves performance over simple multitasking. In order to reduce computation time, we train these models with the FoV$\times$4 dataset. Finally, we train the two best models, \resuneta d6 cmtsk and \resuneta d7v2 cmtsk on the FoV$\times$1 dataset, in order to see if there are performance differences due to different Field of Views as well as compare our models with the results from other authors. }

\subsection{Performance of \resuneta on the FoV$\times$4 dataset}

\resuneta d6  (i.e. the model with no multi-task output) shows competitive performance in all classes (Table \ref{resuneta_between_comparison}).  
The worst result (excluding the class ``Background'') for this model comes in the class ``Trees``, where it seems that \resuneta d6 systematically under segments the area close to their  boundary. \textcolor{black}{This is \emph{partially} owed to the fact that we reduced the size of the original image, and fine details  required for the detailed extent of trees cannot be identified by the algorithm}. In fact, even for a human, the annotated boundaries of trees are not always clear (e.g. see Fig. \ref{resuneta_potsdam_2_13}). 
The \resuneta d6 cmtsk model provides a significant performance boost over the single task \resuneta d6 model, for the classes ``Bulding'', ``LowVeg'' and ``Tree''. 
In these classes it also outperforms the deeper models d7\textcolor{black}{v1} (which, however, do not include the \texttt{PSPPooling} layer at the end of the encoder).  This is  due to the explicit requirement for the algorithm to reconstruct also the boundaries and the distance map and use them to further refine the segmentation mask. As a result, the algorithm gains a ``better understanding'' of the fine details of objects, even if in some cases it is difficult for humans to clearly identify their boundaries.

The \resuneta d7\textcolor{black}{v1} cmtsk model demonstrates slightly increased performance over all of the tested models 
(Table \ref{resuneta_between_comparison}, although differences are marginal for the FoV$\times$4 dataset, and vary between classes).  In addition, there are some annotation errors to the dataset that eventually prove to be an upper bound to the performance.  In Fig. \ref{resuneta_potsdam_2_13} we give an example of inference on 256x256 patches of images on unseen test data. 
In Fig. \ref{resuneta_inference_all_demo} we provide an example of the inference performed by \resuneta d7\textcolor{black}{v1} cmtsk for all the predictive tasks (boundary, distance transform, segmentation, and identity reconstruction). In all rows, the left column corresponds to the same ground truth image. In the first row, from left to right: input image, ground truth segmentation mask, inference segmentation mask. Second row, middle and right: ground truth boundary and inference heat map of the confidence of the algorithm for characterizing pixels as boundaries. The more faint the boundaries are, the less confident is the algorithm for their characterization as boundaries. Third row, middle and right: distance map and inferred distance map. Last row, middle: reconstructed image in HSV space. Right image: average error over all channels between the original RGB image and the reconstructed one. The reconstruction is excellent suggesting that the Tanimoto loss can be used for identity mappings, whenever these are required (as a means of regularization or for Generative Adversarial Networks training \citep{NIPS2014_5423}, e.g. \citet{DBLP:journals/corr/ZhuPIE17}). 

Finally, in Table \ref{resuneta_between_comparison}, we provide a relative comparison between models trained in the FoV$\times$4 and FoV$\times$1  versions of the datasets. Clearly, there is a performance boost when using the higher resolution dataset (FoV$\times$1) for the classes that require finer details. However, for the class ``Building'' the score is actually better with the wider Field of View (FoV$\times$4, model d6 cmtsk) dataset.

\begin{table*}
\footnotesize
\caption{Potsdam comparison of results (based on per class F1 score) with other authors. Best values are marked with bold, second best values are underlined, third best values are in square brackets. \textcolor{black}{Models trained with FoV$\times$1 were trained on 256x256 patches extracted from the original resolution images.}}
\label{resuneta_comparison}
\begin{center}
\begin{tabular}{ l  c  c  c  c  c  c  c  }\hline
Methods  	&ImSurface	&Building &LowVeg &Tree &Car &Avg. F1 & OA\\ \hline
UZ\_1 \citep{volpi2017dense} & 89.3 & 95.4 & 81.8 & 80.5 & 86.5 & 86.7 & 85.8 \\
RIT\_L7  \citep{liu2017dense}  & 91.2 & 94.6 & 85.1 & 85.1 & 92.8 & 89.8 & 88.4 \\
RIT\_4  \citep{rs10091429}  &92.6	&97.0	&86.9	&87.4	&95.2	& 91.8  &90.3\\
DST\_5 \citep{DBLP:journals/corr/Sherrah16}  & 92.5 & 96.4 & 86.7 & \underline{88.8} & 94.7 &  91.7 & 90.3 \\
CAS\_Y3 \citep{ISPRS}  & 92.2 & 95.7 & 87.2 & 87.6 & 95.6 & 91.7 & 90.1 \\
CASIA2 \citep{liu2018semantic} & \underline{93.3}	&\underline{97.0}	&[87.7]	&[88.4]	&\underline{96.2} &  \underline{92.5}	& \underline{91.1} \\
DPN\_MFFL \citep{s18113774} & 92.4 & [96.4] & \underline{87.8} & 88.0 & 95.7 & 92.1 & 90.4 \\
HSN+OI+WBP \citep{rs9060522} & 91.8 & 95.7 & 84.4 & 79.6 & 88.3  & 87.9 & 89.4\\[2pt]
\hline \hline
\resuneta d6 cmtsk (FoV$\times$1) & [93.0] & \textbf{97.2} & 87.5 & [88.4] & [96.1] & [92.4] & [91.0]\\ 
\resuneta d7v2 cmtsk (FoV$\times$1) & \textbf{93.5} & \textbf{97.2} & \textbf{88.2} & \textbf{89.2} & \textbf{96.4} & \textbf{92.9} & \textbf{91.5}\\\hline 
\end{tabular}
\end{center}
\end{table*}

\subsection{Comparison with other modeling frameworks}

\begin{figure}[h!]
\centering
\includegraphics[width=\columnwidth]{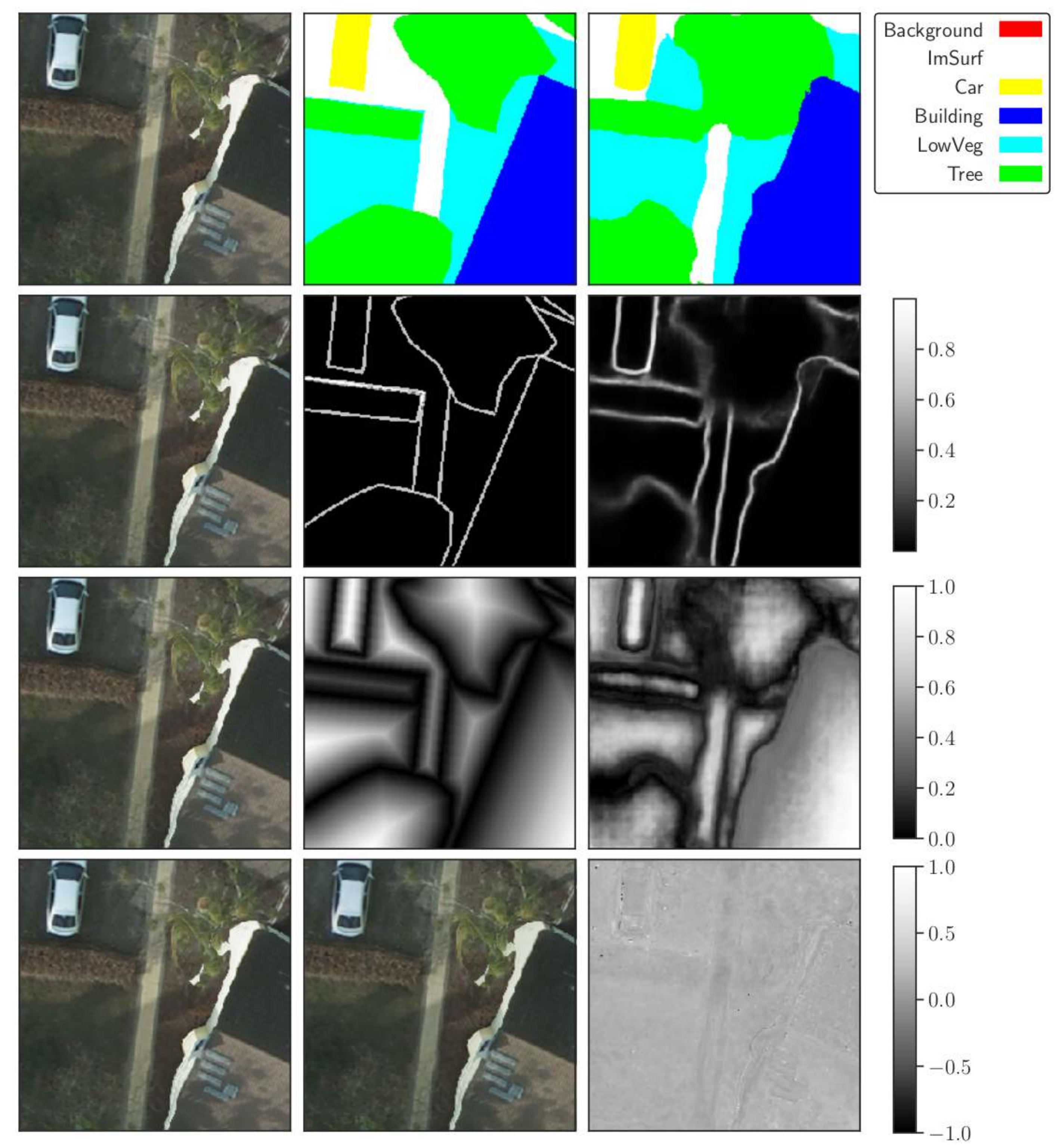}
\caption{ Same as Fig. \ref{resuneta_inference_all_demo} for the \resuneta d7\textcolor{black}{v2} cmtsk trained on the \textcolor{black}{FoV$\times$1 dataset \textcolor{black}{(256$\times$256 image patches, ground sampling distance 5cm)}. It is clear that finer details are present especially for the class ``Trees'' and ``LowVeg'', that improve the performance of the algorithm over the FoV$\times$4 dataset.}}
\label{resuneta_inference_allFoV1_demo}
\end{figure}

\textcolor{black}{In this section, we compare the performance of the \resuneta modeling framework with a representative sample of (peer reviewed) alternative convolutional neural network models. For this comparison we evaluate the models trained on the FoV$\times$1 dataset. The modeling frameworks we compare against \resuneta have published results on the ISPRS website. These are: UZ\_1 \citep{volpi2017dense}, RIT\_L7  \citep{liu2017dense}, RIT\_4  \citep{rs10091429}, DST\_5 \citep{DBLP:journals/corr/Sherrah16}, CAS\_Y3 \citep{ISPRS}, CASIA2 \citep{liu2018semantic}, DPN\_MFFL \citep{s18113774}, and HSN+OI+WBP \citep{rs9060522}. 
To the best of our knowledge, at the time of writing this manuscript, these consist of the best performing models in the competition. 
For comparison, we provide the F1-score per class over all test tiles, the average F1-score over all classes over all test tiles, and the overall accuracy. Note that the average F1-score was calculated using all classes except the ``Background'' class. The overall accuracy, for \resuneta, was calculated including the ``Background'' category.}

 In Table \ref{resuneta_comparison} we provide the comparative results, per class, as well as the average F1 score and overall accuracy for the \resuneta d6 cmtsk and \resuneta d7v2 cmtsk  models as well as  results from other authors. 
 \resuneta d6 performs very well in accordance with other state of the art modeling frameworks, and it ranks overall 3rd (average F1). It should be stressed that for the majority of the results, the performance differences are marginal. 
Going deeper, the \resuneta d7v2 model 
rank 1st among the representative sample of competing models, in all classes, thus clearly demonstrating the improvement over the state of the art.  In Table \ref{ConfMatrix_ALL} we provide the confusion matrix, over all test tiles, for this particular model. 

It should be  noted that some of the contributors (\eg CASIA2, RIT\_4, DST\_5) in the ISPRS competition used networks with pre-trained weights on external large data sets \citep[e.g. ImageNet,][]{imagenet_cvpr09} and fine-tuning, i.e. a methodology called transfer learning \citep[][see also \cite{7301382}, \citet{DBLP:journals/corr/XieJBLE15} for remote sensing applications]{Pan:2010:STL:1850483.1850545}. In particular, CASIA2, that has the 2nd highest overall score, used as a basis a  state of the art pre-trained ResNet101 \citep{DBLP:journals/corr/HeZR016} network.  In contrast, \resuneta was trained from random weights initialization only on the ISPRS Potsdam data set. Although it has been demonstrated that such a strategy does not influence the final performance, i.e. it is possible to achieve the same performance without pre-trained weights \citep{DBLP:journals/corr/abs-1811-08883}, this comes at the expense of a very long training time.

\begin{table*}
\footnotesize
\caption{Potsdam summary confusion matrix over all test tiles for ground truth masks that do not include the boundary. The results correspond to the best model, \resunetacmtsk d7v2, trained on the FoV$\times$1 dataset. The overall accuracy achieved is \bf{91.5\%}}. 
\label{ConfMatrix_ALL}
\begin{center}
\begin{tabular}{ | l | c | c | c | c | c | c | }\hline
\diagbox{Predicted}{Reference} 
	&ImSurface
	&Building
	&LowVeg
	&Tree
	&Car
	&Clutter/Background\\ \hline
ImSurface
&     \bf{0.9478} &    0.0085 &  0.0247 &  0.0117 &  0.0002 &              0.0071
\\ 
Building
&     0.0115 &    \bf{0.9765} &  0.0041 &  0.0025 &  0.0001 &              0.0053
\\ 
LowVeg
&     0.0317 &    0.0057 &  \bf{0.9000} &  0.0532 &  0.0000 &              0.0095 
\\ 
Tree
&     0.0223 &    0.0036 &  0.0894 &  \bf{0.8807} &  0.0016 &              0.0024 
\\ 
Car
&     0.0070 &    0.0016 &  0.0002 &  0.0091 &  \bf{0.9735} &              0.0086 
\\ 
Clutter/Background
&     0.2809 &    0.0844 &  0.1200 &  0.0172 &  0.0090 &              \bf{0.4885}
\\\hline \hline
Precision/Correctness
& 0.9220 & 0.9679 &0.8640 &0.9030 & 0.9538 & 0.7742\\ 
Recall/Completeness
& 0.9478 & 0.9765 & 0.9000 & 0.8807 & 0.9735 &0.4885
\\ \hline \hline
F1 & 0.9347 & 0.9722& 0.8816 & 0.8917 & 0.9635 & 0.5990
\\ \hline
\end{tabular}
\end{center}
\end{table*}

\begin{figure*}
\centering
\includegraphics[width=\textwidth,trim=3.5cm 3.5cm .5cm .5cm,clip]{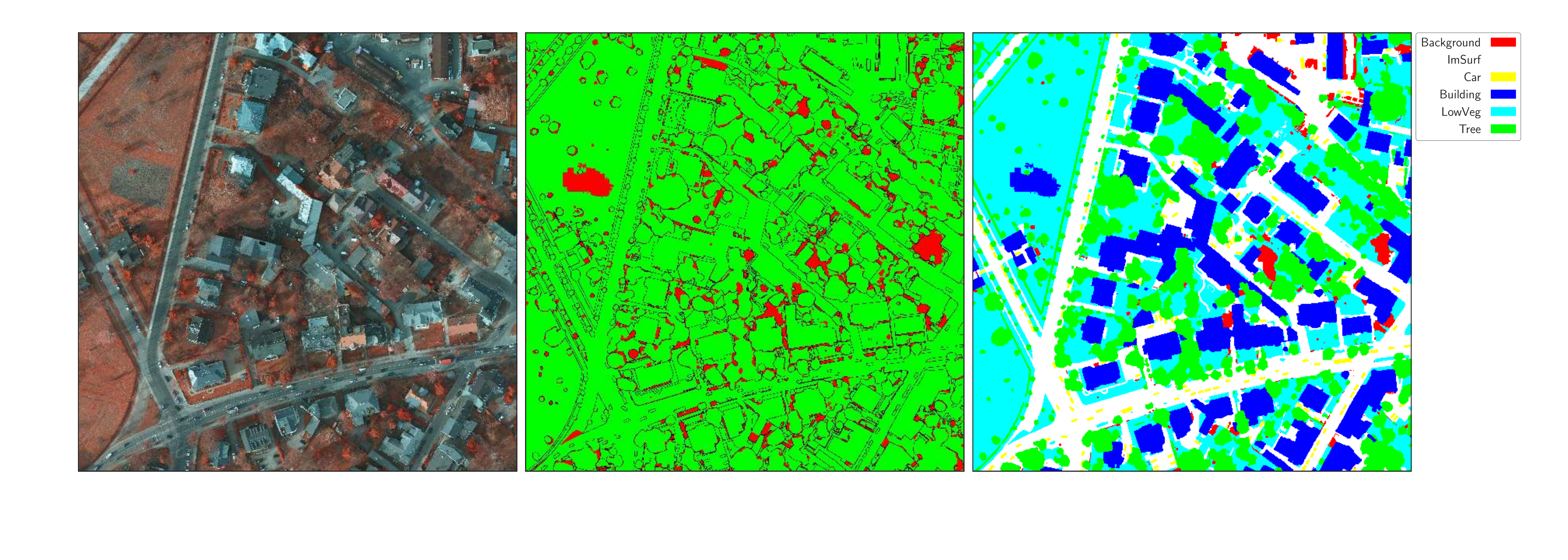}
\includegraphics[width=\textwidth,trim=3.5cm 3.5cm .5cm .5cm,clip]{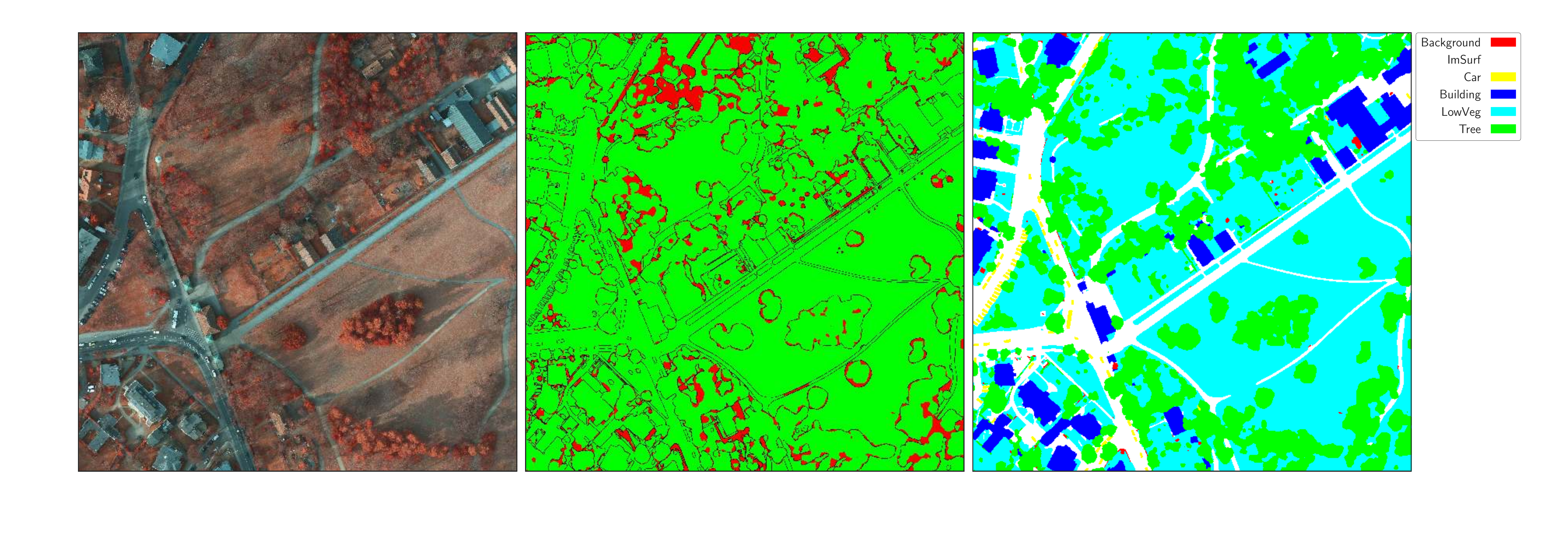}
\includegraphics[width=\textwidth,trim=3.5cm 3.5cm .5cm .5cm,clip]{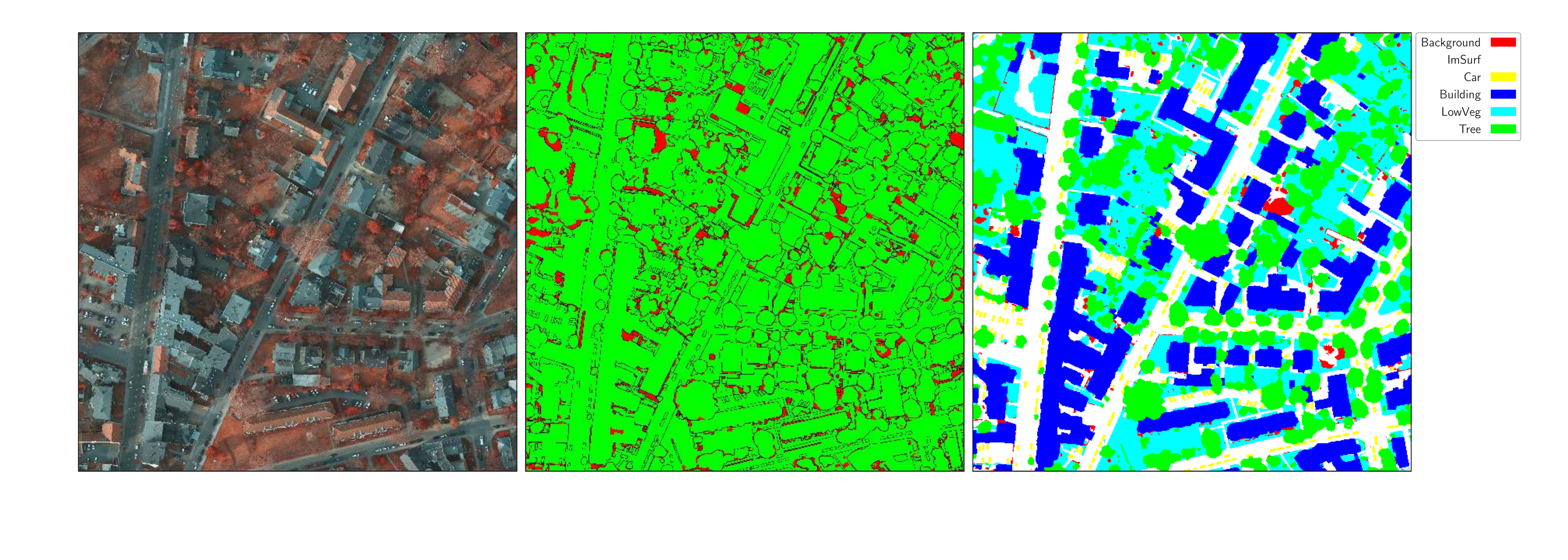}
\includegraphics[width=\textwidth,trim=3.5cm 3.5cm .5cm .5cm,clip]{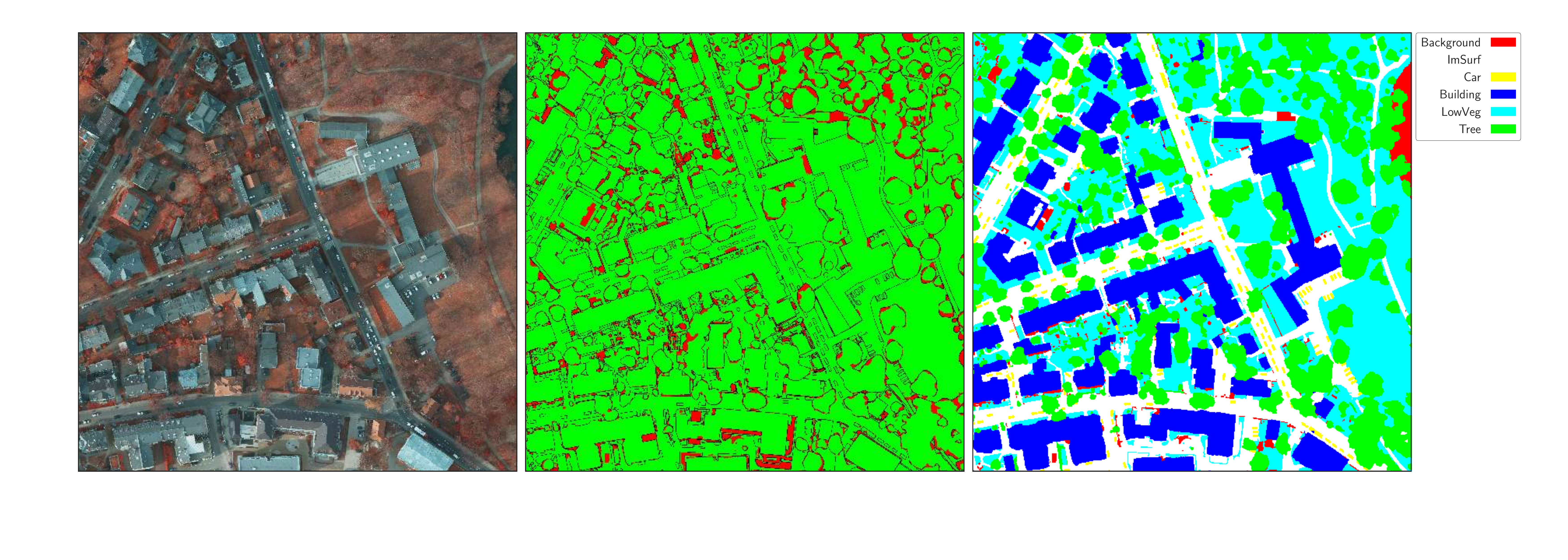}
\caption{\resuneta best model results for tiles 2-13, 2-14, 3-13 and 3-14.  From left to right, input image, difference between ground truth and predictions, inference map. \textcolor{black}{Image resolution:6k$\times$6k, ground sampling distance of 5cm}.}
\label{resuneta_potsdam_sample_error}
\end{figure*}

\textcolor{black}{To visualize the performance of \resuneta, we generated error maps that indicate incorrect (correct) classification in red (green). All summary statistics and error maps were created using the software provided on the ISPRS competition website. For all of our inference results, we used the ground truth masks with eroded boundaries as suggested by the curators of the ISPRS Potsdam data set~\citep{ISPRS}. This allows interested readers to have a clear picture of the strengths and weaknesses of our algorithm in comparison with online published results\footnote{For comparison, competition results can be found \href{http://www2.isprs.org/commissions/comm2/wg4/potsdam-2d-semantic-labeling.html}{online}.}}.
In Fig. \ref{resuneta_potsdam_sample_error} we provide the input image (left column), the error map between the inferred and ground truth masks (middle column) and the inference (right column) for a sample of four test tiles. 
In \ref{resuneta_inf_results} we present the evaluation results for the rest of the test TOP tiles, per class. In all of these figures, for each row, from left to right: original image tile, error map and inference using our best model (\resunetacmtsk d7\textcolor{black}{v2}).

\section{Conclusions}

In this work, we present a new deep learning modeling framework, for semantic segmentation of high resolution aerial images. The framework consists of a novel multitasking deep learning architecture for semantic segmentation and a new variant of the Dice loss that we term Tanimoto.

Our deep learning architecture, \texttt{ResUNet-a}, is based on the encoder/decoder paradigm, where standard convolutions are replaced with ResNet units that contain multiple in parallel atrous convolutions.  Pyramid scene parsing pooling is included in the middle and end of the network. The best performant variant of our models are conditioned multitasking models which predict among with the segmentation mask also the boundaries of the various classes, the distance transform (that provides information for the topological connectivity of the objects) as well as the identity reconstruction of the input image. The additionally inferred tasks, are re-used internally into the network before the final segmentation mask is produced. That is, the final segmentation mask is conditioned on the inference result of the boundaries of the objects as well as the distance transform of their segmentation mask. We show experimentally that the conditioned multitasking improves the performance of the  inferred semantic segmentation classes. 
The ground truth labels that are used during training for the boundaries, as well as the distance transform, can be both calculated very easily from the ground truth segmentation mask using standard computer vision software (\textsc{OpenCV}, see Section \ref{bound_dist_estimate} for a \textsc{Python} implementation). 

We analyze the performance of various flavours of the Dice loss and introduce a novel variant of this as a loss function, the Tanimoto loss. This loss can also be used for regression problems. This is an appealing property that makes this loss useful for the case of multitasking problems in that it results in balanced gradients for all tasks during training. 
We show experimentally that the Tanimoto loss speeds up the training convergence and behaves well under the presence of heavily imbalanced data sets.

The performance of our framework is evaluated on the 2D semantic segmentation \cite{ISPRS} Potsdam data set. Our best model, \resuneta d7v2 achieves top rank performance in comparison with other published results (Table \ref{resuneta_comparison}) and demonstrates a clear improvement over the state of the art. 
The combination of \resuneta conditioned multitasking with the proposed loss function is a reliable solution for performant semantic segmentation tasks.

\section*{Acknowledgments}
The authors acknowledge the support of the Scientific Computing team of CSIRO, and in particular Peter H. Campbell and Ondrej Hlinka.  Their contribution was substantial in overcoming many technical difficulties of distributed \textsc{GPU} computing. The authors are also grateful to John Taylor for his help in understanding and implementing distributed optimization using \textsc{Horovod} \citep{sergeev2018horovod}.  The authors acknowledge  the support of the \textsc{mxnet} community, and in particular Thomas Delteil, Sina Afrooze and Thom Lane. 
The authors  acknowledge the provision of the Potsdam ISPRS dataset by BSF Swissphoto\footnote{\href{http://www.bsf-swissphoto.com/unternehmen/ueber\_uns\_bsf.}{http://www.bsf-swissphoto.com/unternehmen/ueber\_uns\_bsf.}}. \textcolor{black}{The authors acknowledge the contribution of the anonymous referees, whos questions helped to improve the quality of the manuscript.}

\section*{References}
\bibliography{AI_BIB}

\appendix 

\section{\textcolor{black}{Software implementation and training characteristics}}
\label{resuneta_babysitting}

\begin{figure}
\centering
\includegraphics[width=\columnwidth]{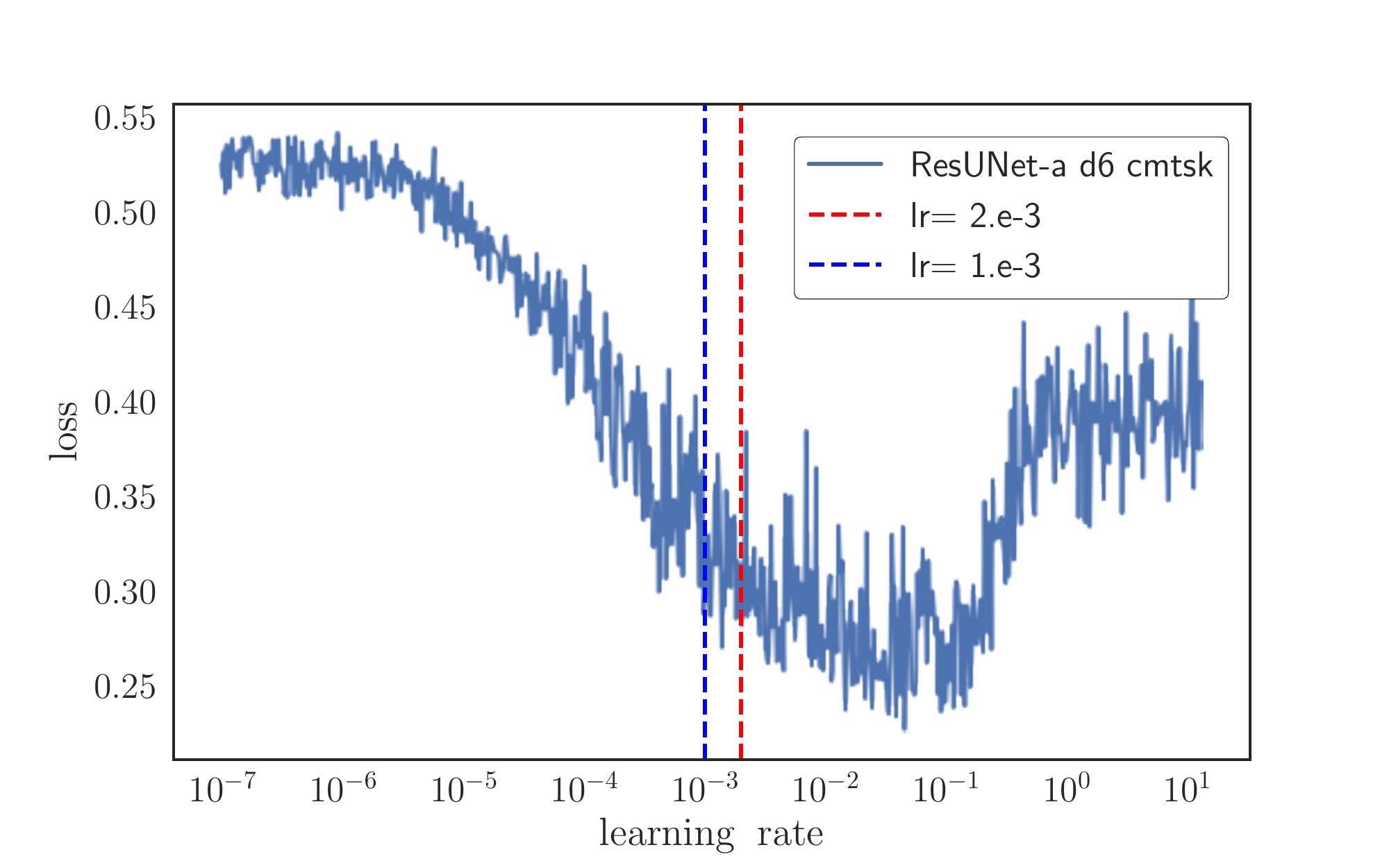}
\caption{Learning rate finder process for the FoV$\times$1 dataset. The model used was \resuneta d6 cmtsk. For this particular training profile, our learning rate choice was 0.001, although a little higher values are possible (see \citealt{DBLP:journals/corr/abs-1803-09820} for details.}
\label{resuneta_lr_finder}
\end{figure}

\resuneta was built and trained using the \textsc{mxnet} deep learning library \citep{chen2015mxnet}, under the \textsc{GLUON} API. Each of the models trained on the FoV$\times$4 dataset  was trained with a batch size of 256 on a single node containing 4 NVIDIA Tesla P100 GPUs  in CSIRO HPC facilities. Due to the complexity of the network, the batch size in a single GPU iteration cannot be made larger than $\sim$ 10 (per GPU). In order to increase the batch size we used manual gradient aggregation\footnote{A a short tutorial on manual gradient aggregation with the \textsc{gluon} API in the \textsc{mxnet} framework can be found \href{https://www.linkedin.com/pulse/increasing-batch-size-under-gpu-memory-limitations-diakogiannis/}{online}.}. 
For the models trained on the FoV$\times$1 dataset we used a batch size of 480 
in order to speed up the computation. 
These  were trained in a  distributed scheme, using the ring 
allreduce algorithm, and in particular it's implementation on \textsc{Horovod} \citep{sergeev2018horovod} for the \textsc{mxnet} \citep{chen2015mxnet} deep learning library. The optimal learning rate for all runs was set by the methodology developed in \cite{DBLP:journals/corr/abs-1803-09820}. In particular, by monitoring the loss error during training for a continuously increasing learning rate, starting from a very low value. An example is shown in Fig. \ref{resuneta_lr_finder}: The optimal learning rate is approximately the point of steepest decent of the loss functions.  This process was complete in approximately 1 epoch and it can be applied in a distributed scheme as well. We found it more useful than the linear learning rate scaling that is used for large batch size \citep{DBLP:journals/corr/GoyalDGNWKTJH17}  in distributed optimization.

For all models , we used the Adam \citep{DBLP:journals/corr/KingmaB14} optimizer, with an initial learning rate of 0.001 (initial learning rate can also be set higher for this dataset, see Fig. \ref{resuneta_lr_finder}), momentum parameters $(\beta_1,\beta_2)=(0.9,0.999)$. 
The learning rate was reduced by an order of magnitude  whenever the validation loss stopped decreasing.  Overall we reduced the learning rate 3 times. We have also experimented  with smaller batch sizes. In particular, with a batch size of 32, the training is unstable. This is owed mainly to the fact that we used 4 GPUs for training, therefore the batch size per GPU is 8, and this is not sufficient for the Batch Normalization layers that use only the data per GPU for the estimation of running means of their parameters. When we experimented with synchronized Batch Normalization layers  \citep{DBLP:journals/corr/IoffeS15,Zhang_2018_CVPR}, this increased the stability of the training dramatically even with a batch size as small as 32. However, due to the GPU synchronization, this was a slow operation that proved to be impractical for our purposes. 

A software implementation for the \resuneta models that relate to this work can be found on github\footnote{\href{https://github.com/feevos/resuneta}{https://github.com/feevos/resuneta}}.

\section{Boundary and  distance transform from segmentation mask}
\label{bound_dist_estimate}
The boundaries and distance transform can be estimated efficiently from the segmentation ground truth mask by the  python software routines listed here. The input \texttt{labels} is a binary image, with 1 designating on class and 0 off class pixels. The shape of the \texttt{labels} is two dimensional (i.e. it is a single channel image, of shape \texttt{(Height},\texttt{Width}) - no channel dimension). 
In a multiclass context the segmentation mask must be provided in one-hot encoding and applied iteratively per channel.   


\newpage
\begin{python}
import cv2
import numpy as np

def get_boundary(label, kernel_size = (3,3)):
    tlabel = label.astype(np.uint8)
    temp = cv2.Canny(tlabel,0,1)
    tlabel = cv2.dilate(
              temp,
              cv2.getStructuringElement(
              cv2.MORPH_CROSS,
              kernel_size),
              iterations = 1)
    tlabel = tlabel.astype(np.float32)
    tlabel /= 255.
    return tlabel

def get_distance(label):
    tlabel = label.astype(np.uint8)
    dist = cv2.distanceTransform(tlabel,     
                                 cv2.DIST_L2,
                                 0)
    dist = cv2.normalize(dist,
                         dist,
                         0, 1.0,        
                         cv2.NORM_MINMAX)
    return dist            
\end{python}

\section{Inference results}
\label{resuneta_inf_results}
In this section, we present classification results and error maps for all the test TOP tiles of the Potsdam \cite{ISPRS} dataset. 

\begin{figure*}
\centering
\includegraphics[width=\textwidth,trim=3.5cm 3.5cm .5cm .5cm,clip]{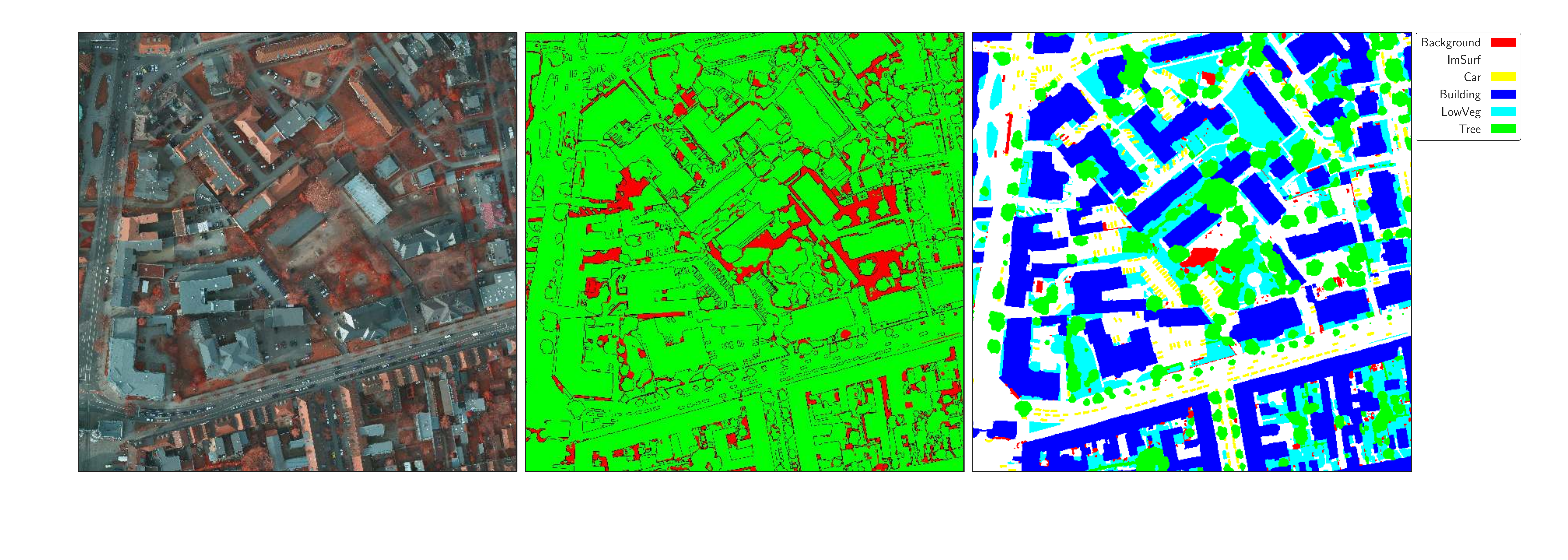}
\includegraphics[width=\textwidth,trim=3.5cm 3.5cm .5cm .5cm,clip]{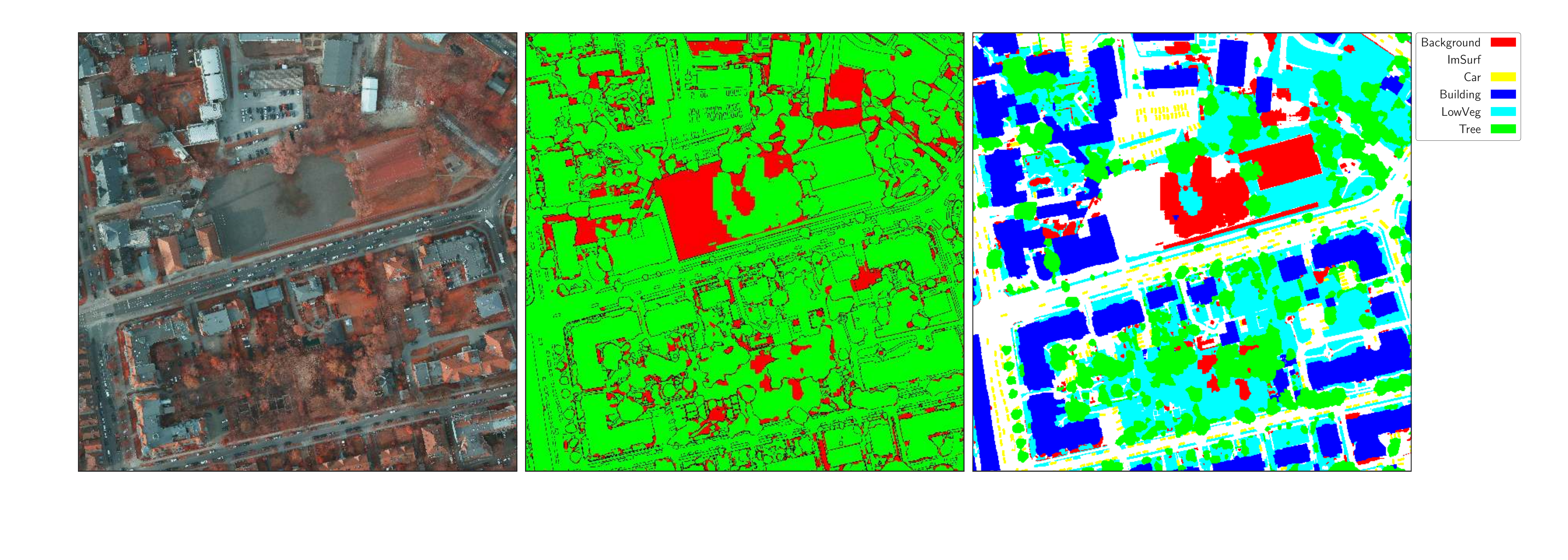}
\includegraphics[width=\textwidth,trim=3.5cm 3.5cm .5cm .5cm,clip]{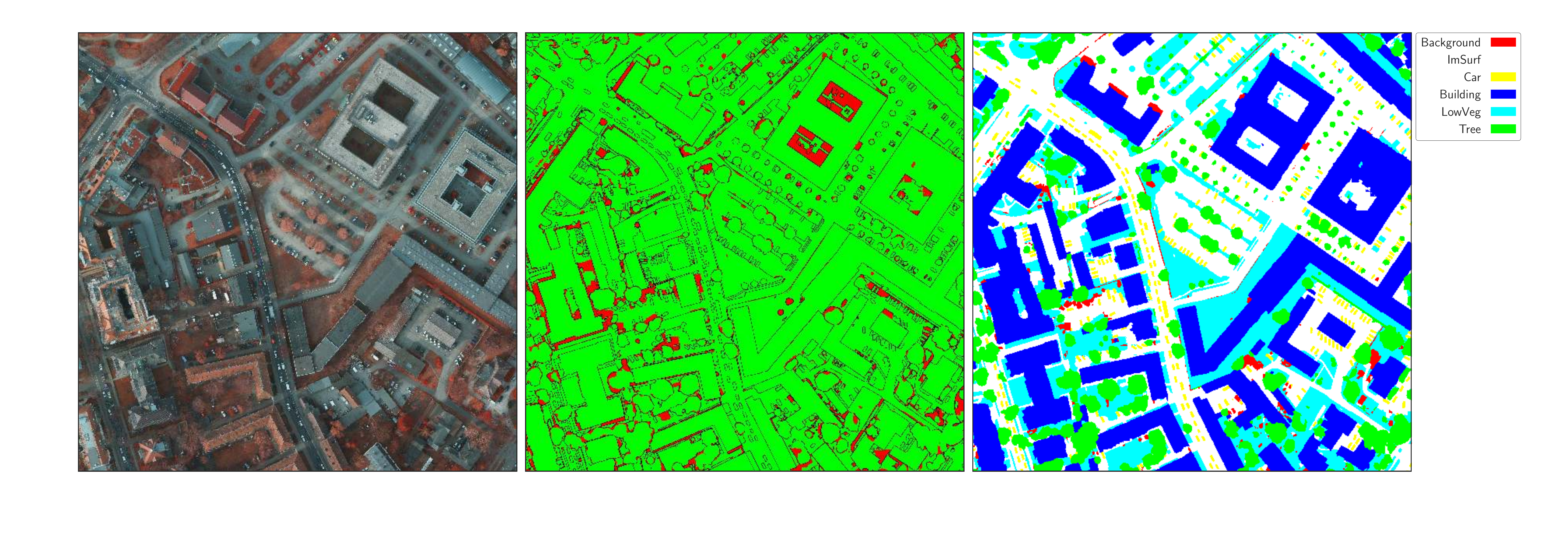}
\includegraphics[width=\textwidth,trim=3.5cm 3.5cm .5cm .5cm,clip]{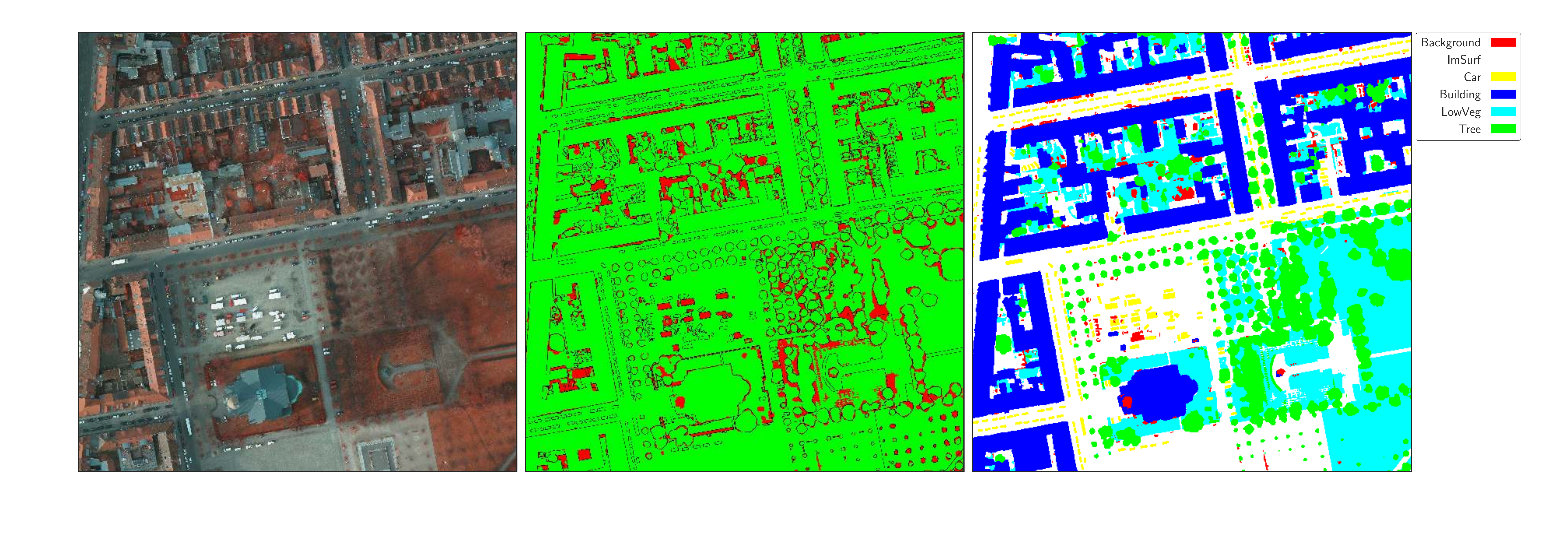}
\caption{\resuneta best model results for tiles 4-13, 4-14, 4-15, 5-13.  From left to right, input image, difference between ground truth and predictions, inference map. \textcolor{black}{Image resolution:6k$\times$6k, ground sampling distance of 5cm}.}
\label{resuneta_potsdam_2_13_3_13}
\end{figure*}

\begin{figure*}
\centering
\includegraphics[width=\textwidth,trim=3.5cm 3.5cm .5cm .5cm,clip]{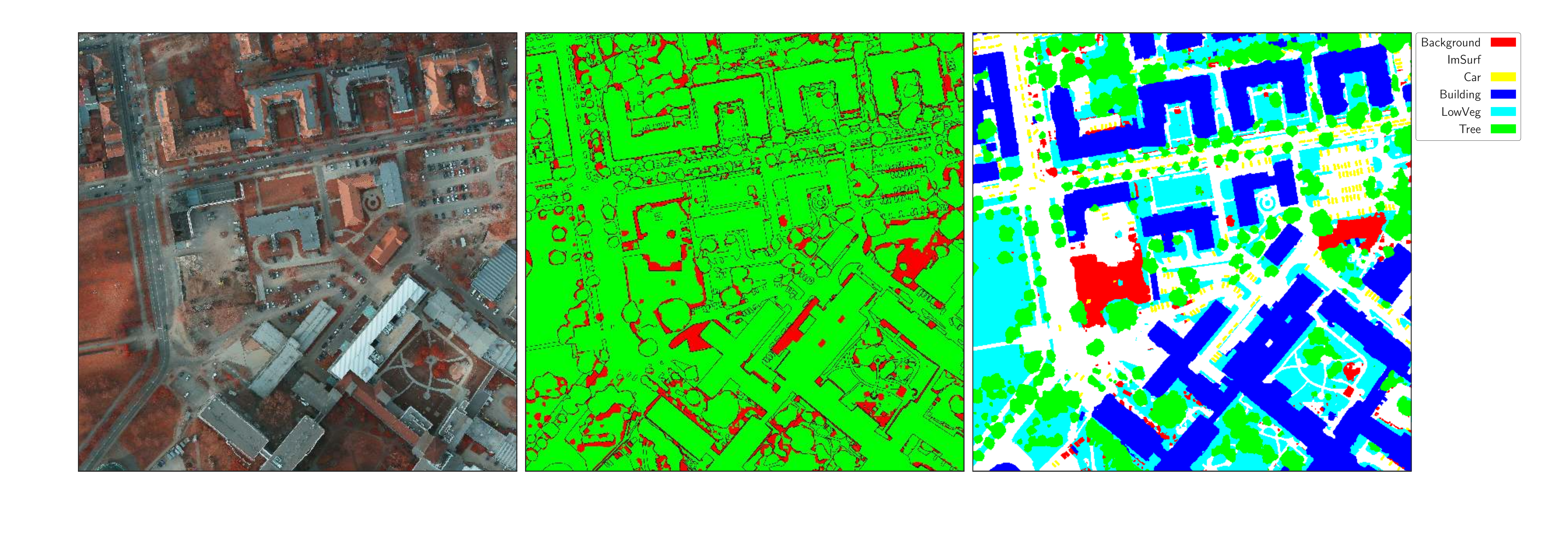}
\includegraphics[width=\textwidth,trim=3.5cm 3.5cm .5cm .5cm,clip]{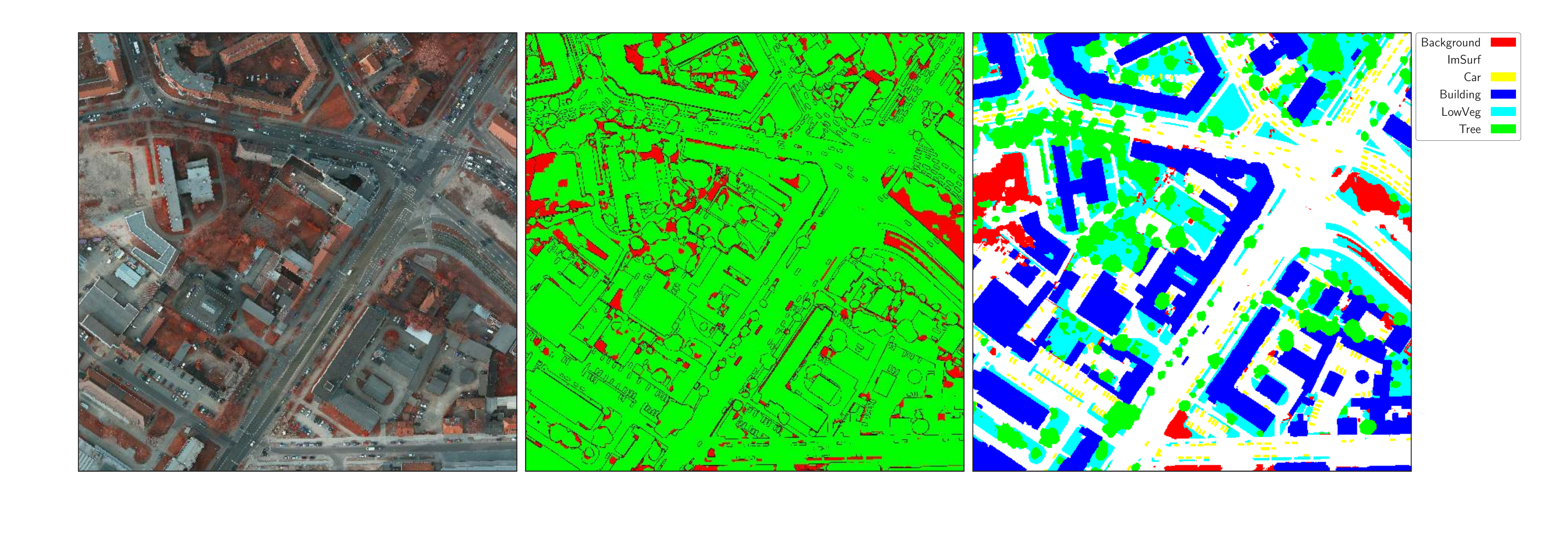}
\includegraphics[width=\textwidth,trim=3.5cm 3.5cm .5cm .5cm,clip]{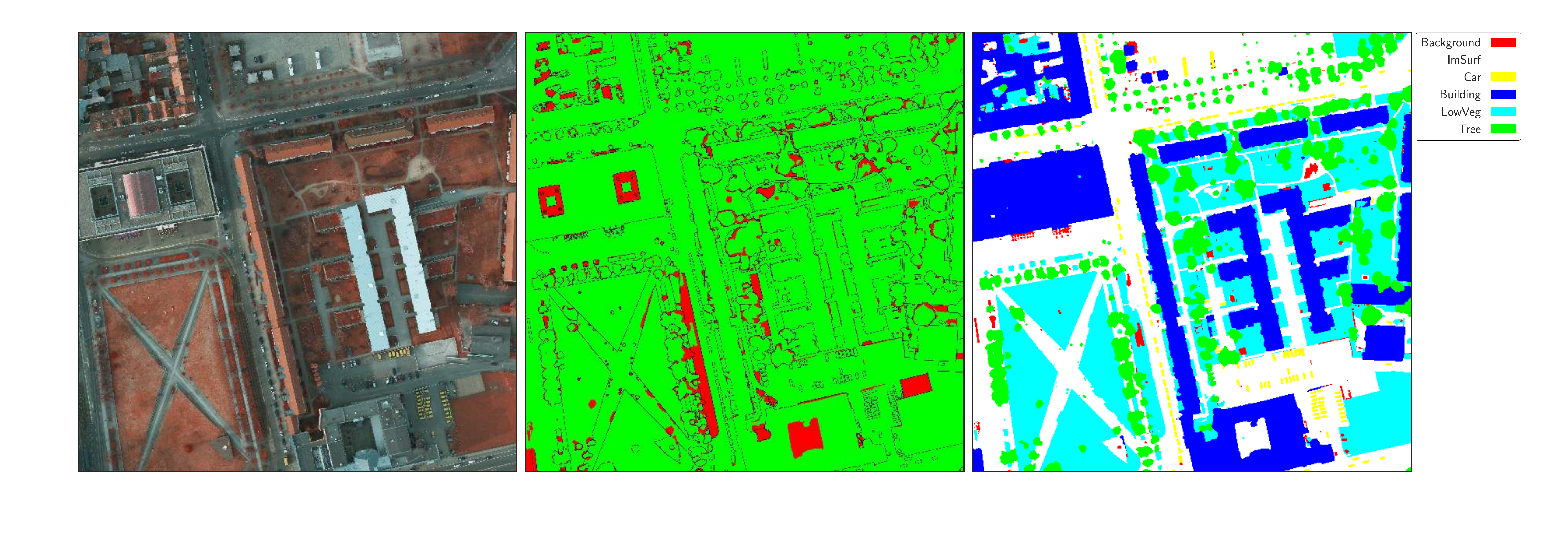}
\includegraphics[width=\textwidth,trim=3.5cm 3.5cm .5cm .5cm,clip]{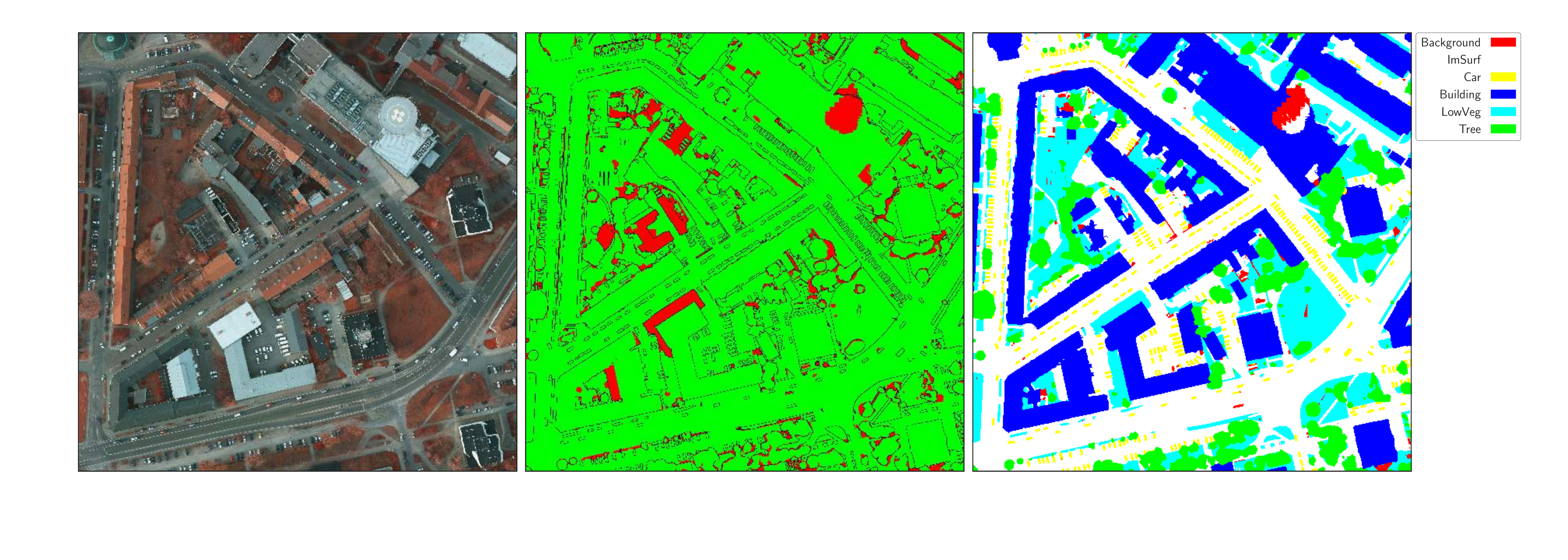}
\caption{As Fig. \ref{resuneta_potsdam_2_13_3_13} for tiles 5-14, 5-15, 6-13, 6-14}
\label{resuneta_potsdam_4_15_5_14}
\end{figure*}

\begin{figure*}
\centering
\includegraphics[width=\textwidth,trim=3.5cm 3.5cm .5cm .5cm,clip]{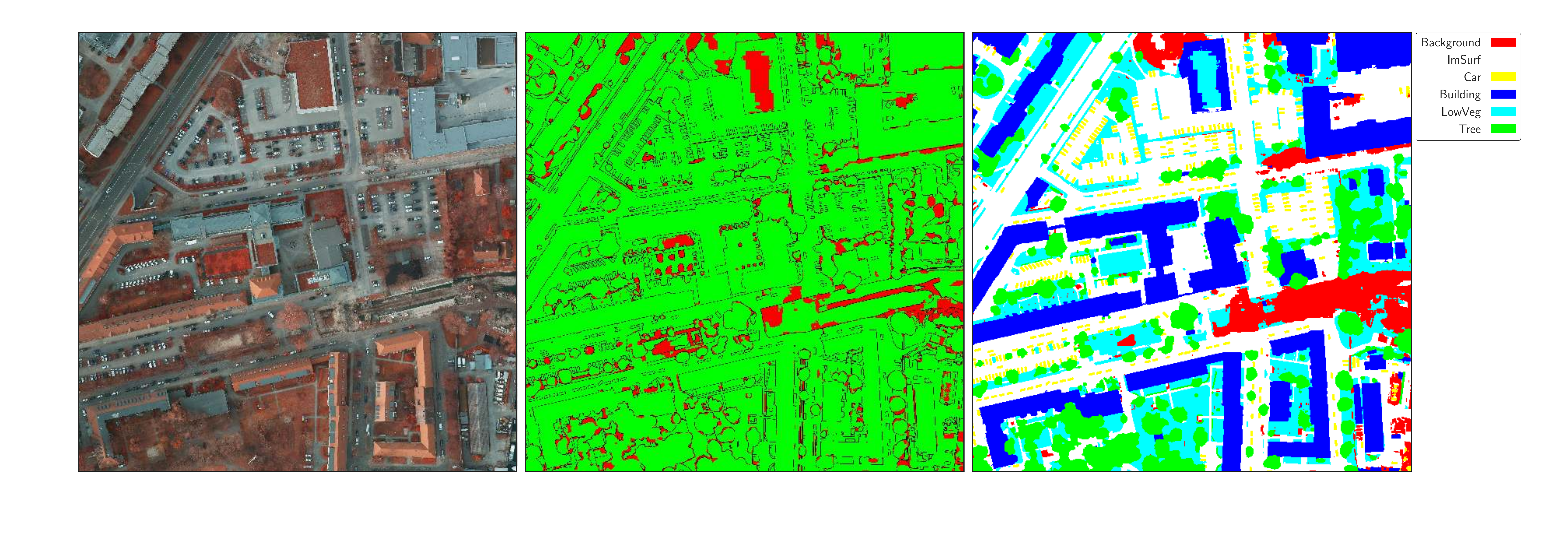}
\includegraphics[width=\textwidth,trim=3.5cm 3.5cm .5cm .5cm,clip]{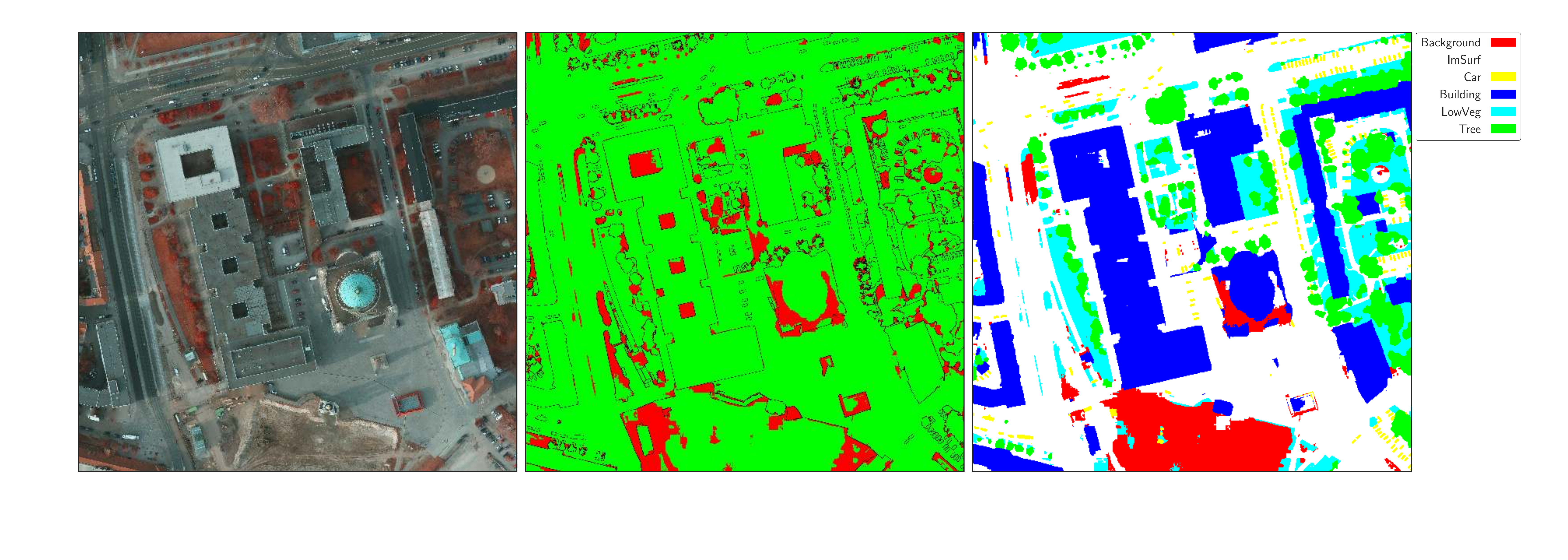}
\caption{As Fig. \ref{resuneta_potsdam_2_13_3_13} for tiles 6-15, 7-13}
\label{resuneta_potsdam_6_15_7_13}
\end{figure*}

\end{document}